\documentclass[11pt]{amsart}

\usepackage[margin=1.25in]{geometry}
\usepackage[colorlinks=true,citecolor=blue,linkcolor=blue]{hyperref}

\usepackage{amsmath, amssymb, amsthm}
\usepackage[dvipsnames]{xcolor}
\usepackage{soul}
\usepackage{graphicx}
\usepackage{wrapfig}
\usepackage{subcaption}
\usepackage{soul,xcolor}
\usepackage{tikz}
\usepackage{todonotes}
\usepackage{verbatim}
\usepackage{float}

\newtheorem{theorem}{Theorem}[section]
\newtheorem{cor}[theorem]{Corollary}

\newtheorem{conj}[theorem]{Conjecture}
\newtheorem{remark}{Remark}
\newtheorem{lem}[theorem]{Lemma}
\newtheorem{defn}[theorem]{Definition}

\DeclareMathOperator{\Ima}{Im}
\DeclareMathOperator{\code}{code}
\newcommand{\R}{\mathbb{R}}

\renewcommand{\P}{\mathbb{P}}
\newcommand{\ImF}{\Ima(F_1)}

\newcommand{\Su}{\mathcal S}
\newcommand{\Hp}[1]{\mathcal H_{#1}}
\newcommand{\halfax}[1]{\mathtt{X}_{#1\ge0}}

\newcommand{\bb}[1]{\mathbb{#1}}
\newcommand{\df}{\dim_{\operatorname{fun}}}

\definecolor{cornflower}{rgb}{0.5067352811380239, 0.4923960137178009, 0.7184954982852749}

\title{Stably Unactivated Neurons in ReLU Neural Networks}
\author[N.\ Brownlowe]{Natalie Brownlowe}
\author[C.R.\ Cornwell]{Christopher R.\ Cornwell}
\author[E.\ Montes]{Ethan Montes}
\author[G.\ Quijano]{Gabriel Quijano}
\author[G.\ Stulman]{Grace Stulman}
\author[N.\ Zhang]{Na Zhang}
\date{\today}

\begin{document}
\begin{abstract}
The choice of architecture of a neural network influences which functions will be realizable by that neural network and, as a result, studying the expressiveness of a chosen architecture has received much attention. In ReLU neural networks, the presence of \textit{stably unactivated} neurons can reduce the network's expressiveness. In this work, we investigate the probability of a neuron in the second hidden layer of such neural networks being stably unactivated when the weights and biases are initialized from symmetric probability distributions. For networks with input dimension $n_0$, we prove that if the first hidden layer has $n_0+1$ neurons then this probability is exactly $\frac{2^{n_0} + 1}{4^{n_0+1}}$, and if the first hidden layer has $n_1$ neurons, $n_1\le n_0$, then the probability is $\frac{1}{2^{n_1+1}}$. Finally, for the case when the first hidden layer has more neurons than $n_0+1$, a conjecture is proposed along with the rationale. Computational evidence is presented to support the conjecture.
\end{abstract}

\maketitle

\section{Introduction and Main Results}\label{sec:Intro}
A feedforward neural network, with selected parameters, has an associated \emph{network function} $F:\bb R^{n_0}\to\bb R^{n_L}$ for some $n_0, n_L \in \bb N$.  This associated function takes as input a data point ${\bf x}$ with $n_0$ \emph{features} (i.e., ${\bf x}\in \bb R^{n_0}$) and, commonly, the output is used for network \emph{prediction} on how to classify ${\bf x}$ into one of $n_L$ possible classes. Throughout this paper, we exclusively consider \emph{ReLU networks} that use activation function $\sigma(x) = \max\{0, x\}$ and have \emph{fully-connected} layers (see Section \ref{sec:prelim}). We study the occurrence of a phenomenon for the network, the existence of what are called \emph{stably unactivated neurons} (see Definition \ref{defn:stably-unactivated}), which influence not only what the network function may be, but the degree to which it can be altered via ``small updates'' to the parameters. 

Neural networks are described, in part, by an \emph{architecture} that consists of a list of natural numbers. Feedforward neural networks with $L$ fully-connected layers and architecture $(n_0, n_1, \ldots, n_L)$ are parameterized by $\theta \in \bb R^D$, where $D = \sum_{i=1}^Ln_{i}(n_{i-1}+1)$.  For each $\theta\in\bb R^D$,  denote the network function for that neural network by $F^\theta$.  Due to symmetries that arise from permuting neurons (coordinates) within a layer, it is well-known that the mapping $\theta\mapsto F^\theta$, from the parameter space $\bb R^D$ to the space of functions realizable by this architecture, is not one-to-one.  Moreover, as discussed in \cite{BuiLampert2020} and \cite{KordingRolnick2020}, for ReLU networks the mapping $\theta\mapsto F^\theta$ is even farther from being one-to-one (provided $L\ge 2$), since it is possible to perform scaling/inverse-scaling on the weights and the bias connected to a neuron in a hidden layer without producing any change to $F^\theta$.

In order to further study the relationship between the parameter space of a given network architecture and the space of realizable functions, the \emph{functional dimension} of $\theta$, denoted $\df(\theta)$, was introduced in \cite{GLMW2022}. As stated in \cite{GLMW2022}, intuitively speaking, $\df(\theta)$ is the number of degrees of freedom within the space of realizable functions that are attainable by arbitrarily small perturbations of $\theta$.
Given $\theta_0\in\bb R^D$, the larger that $\df(\theta_0)$ is, the less ``redundancy'' the mapping $\theta\mapsto F^\theta$ exhibits near $\theta_0$. 
Additionally, when training a ReLU network with gradient descent, updates are restricted to a subspace of the tangent space of $\bb R^D$ at $\theta_0$, which at most has dimension $\df(\theta_0)$.

Interestingly, it was found in \cite{GLMW2022} that, with a fixed architecture, the value of $\df(\theta)$ varies across parameter space. Hence, initialization, or a round of training, might produce a $\theta$ with relatively small functional dimension. As a consequence, the freedom to alter the network function within the space of realizable functions would then be significantly restricted, affecting the ability to train the neural network model effectively.

A consequence of the fact that scaling/inverse scaling on any neuron of a hidden layer leaves the network function unchanged is that the functional dimension of a parameter in $\bb R^D$ is always strictly less than $D$ (provided that there is at least one hidden layer). Using this observation, given an architecture $(n_0, n_1,\ldots, n_L)$ and $D$, as defined above, a tight upper bound on the functional dimension of $\theta\in\bb R^D$ was found in \cite[Theorem 7.1]{GLMW2022}, 
	\begin{equation}\label{eqn:dimfun-upperbound}
	\df(\theta) \le n_L + \sum_{i=1}^Ln_in_{i-1}.
	\end{equation}

The upper bound (\ref{eqn:dimfun-upperbound}) is tight in the sense that there are architectures for which there exists a parameter $\theta$ which has functional dimension equal to the right-hand side of (\ref{eqn:dimfun-upperbound}) \cite[Theorem 8.11, for example]{GLMW2022}. However, as is well-known to experts and practitioners, ReLU networks will often contain some neurons in hidden layers that are ``dead,'' or \emph{unactivated}.  If such a neuron of a network, given by parameter $\theta$, is \emph{stably unactivated} (the generic case of being unactivated; see Definition \ref{defn:stably-unactivated}) then it is impossible for $\df(\theta)$ to achieve equality in (\ref{eqn:dimfun-upperbound}) \cite[Theorem 7.3]{GLMW2022}. 

It is not difficult to see that if the bias of a neuron in layer $i$, $i\in\{2,\ldots, L-1\}$, and every weight ``leading to'' that neuron are all negative, then the neuron will be stably unactivated. When the weights and biases are independent and identically distributed random variables with a distribution symmetric about $0$, then the probability that this occurs is $\frac{1}{2^{1+n_{i-1}}}$. As has been observed, experimental results show that the probability of a neuron in a hidden layer being unactivated is significantly larger. 

In this article, as a step in the direction of understanding the functional dimension as a random variable, we study the probability that a given neuron of the network is stably unactivated. The earliest hidden layer where this probability is non-zero is the second layer of the network, at which our results are focused.  As we note after our main results, generally the (unconditional) probability of a neuron in later layers being stably unactivated increases with the number of the layer.  For our results, we make a mild assumption about the weights and biases that comprise the components of $\theta$.
\begin{theorem}
\label{thm:small-n1}
Let $(n_0,n_1,\ldots,n_L)$ be an architecture of a ReLU neural network. Suppose that the weights and biases in each single layer are selected i.i.d.\ from a probability distribution on $\bb R$ that is symmetric about $0$. If $n_1 \leq n_0$, then for any one of the $n_2$ neurons in the second layer, the probability of that neuron being stably unactivated is precisely 
			\[\frac{1}{2^{{n_1}+1}}.\]
    
\end{theorem}

\begin{theorem}
	Let $(n_0,n_1,\ldots,n_L)$ be an architecture of a ReLU neural network. Suppose that the weights and biases in each single layer are selected i.i.d.\ from a probability distribution on $\bb R$ that is symmetric about $0$. If $n_1 = n_0 + 1$, then for any one of the $n_2$ neurons in the second layer, the probability that the neuron is stably unactivated is precisely 
			\[\frac{2^{n_0} + 1}{4^{{n_0}+1}}.\]
\label{thm:main}
\end{theorem}

As mentioned above, there is a condition that is clearly sufficient for a neuron from the second layer to be stably unactivated and that occurs with probability $\frac{1}{2^{n_1+1}}$. By Theorems \ref{thm:small-n1} and \ref{thm:main}, when $n_1 \le n_0+1$ and $n_1$ is sufficiently large, this probability is either equal or approximately equal to the probability that the neuron is stably unactivated. However, this conclusion finely depends on the conditions, including the layer of the neuron in question. 

For example, there is a positive probability that the image of the function for the network truncated at the second layer is compact or, at least, the projection of the image to one of the coordinates is compact. This increases the probability of a stably unactivated neuron. 

\begin{remark}\label{rem:precomposing}
For a neural network with architecture $(n_0,n_1,\ldots,n_L)$ and $\theta$ selected as in the above theorems, let $p(n_0,n_1)$ be the probability in question in those theorems. Additionally, let $m_0,m_1,\ldots,m_{k-1}$ be positive integers and consider a ReLU network with architecture $(m_0,m_1,\ldots,m_{k-1},n_0,\ldots, n_L)$, with the weights and biases in layers $k+1,\ldots,k+L$ selected identically to those in $\theta$. Choosing a neuron in hidden layer $k+2$, the probability that it is stably unactivated is strictly larger than $p(n_0,n_1)$. 
\end{remark}

Somewhat related to Remark \ref{rem:precomposing} is the main result in \cite{Lu2020}.  A careful reading of the proof of that main result indicates the following. Given any fixed $N > 0$, if we impose the constraint that $n_\ell \le N$ for all $n_\ell$ in networks with architecture $(n_0,n_1,\ldots,n_L)$, then the probability that some layer of the network consists of \emph{only} stably unactivated neurons at $\theta$ limits to 1, as $L \to \infty$. (The assumptions on $\theta$ being the same as in Theorem \ref{thm:main}.)

Remark \ref{rem:precomposing} hints at the idea of understanding a neural network as a composition of networks. The possibility of stably unactivated neurons in some layer then makes it relevant to consider architectures where $n_1$ is large, relative to $n_0$. We provide some evidence and rationale that, in this case, the behavior of the probability addressed in Theorems \ref{thm:small-n1} and \ref{thm:main} is controlled by $n_0$ rather than $n_1$. 

\begin{conj}
\label{conj:large-n1}
Consider architectures $(n_0,n_1,\ldots,n_L)$ with $n_0$ fixed and suppose that the weights and biases in each single layer are selected i.i.d.\ from a probability distribution on $\bb R$ that is symmetric about $0$. Then there exists a constant $c > 0$ such that for $n_1$ sufficiently large, the probability for any one of the $n_2$ neurons 
in the second layer 
being stably unactivated is at least $\frac{c}{4^{n_0}}$.
\end{conj}

The remainder of the paper is organized as follows. After reviewing relevant definitions and background in Section \ref{sec:prelim}, we prove in Section \ref{sec:lemmas} some results on hyperplane arrangements, and we discuss a partition on hyperplanes in $\bb R^{n_1}$ that will be helpful in proving a key lemma, needed for the proof of Theorem \ref{thm:main}. Sections \ref{sec:proof-main} and \ref{sec:configs} contain the proofs of Theorems \ref{thm:small-n1} and \ref{thm:main}, the key Lemma \ref{Lemma: Sum of delta} being proved in Section \ref{sec:configs}. Finally, in Section \ref{sec:general architectures} we discuss Conjecture \ref{conj:large-n1}. In addition to the rationale for the conjecture, we provide some supporting computational evidence.

\section{Notation and Background}\label{sec:prelim}

In this section we introduce terminology and notation that will be needed throughout the paper. Hyperplane arrangements and ideas related to them will be important for the discussion. After reviewing hyperplane arrangements, we discuss ReLU neural networks and their stably unactivated neurons, which are the objects of our main interest.

\subsection{Hyperplane arrangements, regions, and codewords of regions.}
For an integer $n \ge 1$, a hyperplane in $\bb R^{n}$ is the solution set to an affine-linear equation,

\begin{equation}\label{eqn:affine-linear}
\{{\bf x}\in\R^n\ |\ {\bf w}\cdot{\bf x} + b = 0\}, 
\end{equation}
where $b\in\R$ and ${\bf w} \in \R^n$ is a nonzero vector. 

\begin{defn}
\label{defn:arrangement}
 A $\textbf{hyperplane arrangement}$ in $\bb R^n$ is a finite ordered set of hyperplanes in $\R^n$. If $\mathcal{A}$ is a hyperplane arrangement in $\bb R^n$, the connected components of \mbox{$\bb R^n\setminus\bigcup_{H\in\mathcal{A}}H$} are called the \textbf{regions} of $\mathcal{A}$.
\end{defn}

A vector ${\bf w}$, that helps to determine a hyperplane as above, is called a \emph{normal vector} to that hyperplane. While not unique, a choice of normal vector for a hyperplane $H$ provides a natural way to specify which component of $\bb R^n \setminus H$ is called the \emph{positive half-space} $H^+$, and which is the \emph{negative half-space} $H^-$. Specifically, given ${\bf w}\in\bb R^n$ and $b\in\bb R$ so that $H = \{{\bf x}\in\R^n\ |\ {\bf w}\cdot{\bf x} + b = 0\}$, we may set 

\[H^{+} = \{{\bf x}\ |\ {\bf w}\cdot{\bf x} + b >0\};\] \[H^{-} = \{{\bf x}\ |\ {\bf w}\cdot{\bf x} + b <0\}.\]

\begin{defn}
    \label{defn:coorientation}
    Let $\mathcal{A} = \{H_1, H_2,\ldots, H_m\}$ be a hyperplane arrangement in $\R^n$. A \textbf{coorientation} on $\mathcal{A}$ is a specification, for each $i=1,2,\ldots,m$, of which component of $\R^n\setminus H_i$ is the positive half-space {--} the other component being the negative half-space. 
\end{defn}
\begin{remark}
    In this paper, to refer to a hyperplane arrangement with a given coorientation we use bold-face notation. For example, a cooriented hyperplane arrangement may be written as $\textbf{A} = \{H_1, H_2, \ldots, H_m\}$, indicating that the half-spaces $H_1^+, H_2^+,\ldots, H_m^+$ have been specified.
\end{remark}

Given a choice of normal vector for each hyperplane in a hyperplane arrangement $\mathcal{A}$, our previous comments indicate a corresponding natural choice of coorientation on $\mathcal A$. In much of the paper we will assume a coorientation determined by such a choice. Additionally, in the cooriented setting, note that each region of the hyperplane arrangement is determined by knowing which positive half-space it is contained in. 
This information will play a useful role.
Following \cite{ItskovKuninRosen20}, we make the following definition.\footnote{Our codes use vectors in $\{+,-\}^m$ rather than subsets of $\{1,2,\ldots,m\}$ (to indicate containment in a positive halfspace) as done for \emph{hyperplane codes} in \cite{ItskovKuninRosen20}. Other than notational, the only difference in approach is that we care only about points in the union of the regions of the hyperplane arrangement, i.e., in $\bb R^n\setminus\bigcup_{H\in\mathcal A}H$, whereas in \cite{ItskovKuninRosen20} each codeword may apply to part of the boundary of a region also.}

\begin{defn}
    \label{defn:code}
    Let $\textbf{\emph{A}} = \{H_1,H_2,\ldots, H_m\}$ be a hyperplane arrangement in $\bb R^n$, equipped with a coorientation, and let $R$ be a region of $\textbf{\emph{A}}$. There exists a unique $(c_1,c_2,\ldots,c_m)$ in $\{+,-\}^m$ such that $R = \bigcap_{i=1}^m H_i^{c_i}$. We call $(c_1,c_2,\ldots,c_m)$ the \textbf{codeword} of $R$ and will denote it by $\operatorname{code}(R)$. The set of codewords for all regions of $\textbf{\emph{A}}$ is referred to as the \textbf{code} of $\textbf{\emph{A}}$, and is written  as $\operatorname{code}(\textbf{\emph{A}})$. 
\end{defn}

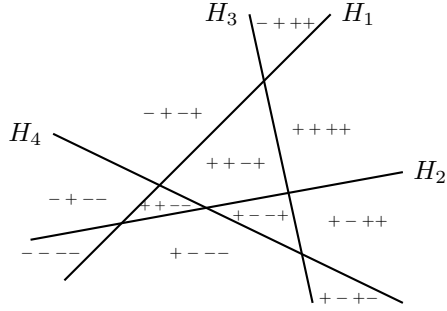
\begin{figure}[H]
    \centering
    \begin{tikzpicture}[scale=0.6]
\draw[black, thick, font=\small] (-1.5,-1.5) -- (4.4,4.4)node[anchor=west]{$H_1$};
\draw[black, thick, font=\small] (-2.25,-0.6) -- (6,0.9)node[anchor=west]{$H_2$};
\draw[black, thick, font=\small] (4,-2) -- (2.6,4.4)node[anchor=east]{$H_3$};
\draw[black, thick, font=\small] (6,-2) -- (-1.75,1.75)node[anchor=east]{$H_4$};
    \node[font=\tiny] at (4.2, 1.85) {\scalebox{0.8}{$++++$}};
    \node[font=\tiny] at (2.3, 1.1) {\scalebox{0.8}{$++-+$}};
    \node[font=\tiny] at (5, -0.2) {\scalebox{0.8}{$+-++$}};
    \node[font=\tiny] at (0.9, 2.2) {\scalebox{0.8}{$-+-+$}};
    \node[font=\tiny] at (1.5, -.9) {\scalebox{0.8}{$+---$}};
    \node[font=\tiny] at (-1.8, -0.9) {\scalebox{0.8}{$----$}};
    \node[font=\tiny] at (3.4, 4.2) {\scalebox{0.8}{$-+++$}};
    \node[font=\tiny] at (2.85, -0.07) {\scalebox{0.75}{$+--+$}};
    \node[font=\tiny] at (-1.2, 0.3) {\scalebox{0.8}{$-+--$}};
    \node[font=\tiny] at (0.75, 0.15) {\scalebox{0.7}{$++--$}};
    \node[font=\tiny] at (4.8, -1.9) {\scalebox{0.8}{$+-+-$}};
\end{tikzpicture}
    \caption{An example generic hyperplane arrangement $\textbf{A}$ in $\bb R^2$ equipped with a coorientation. Codewords of regions are indicated, with $\operatorname{code}(\textbf{A})$ consisting of eleven elements in $\{+,-\}^4$.}
    \label{fig:facet sharing}
\end{figure}

The code of an example hyperplane arrangement with coorientation is illustrated in Figure \ref{fig:facet sharing}. This hyperplane arrangement is \emph{generic} (Definition \ref{defn:generic}). In other work on ReLU networks, some authors have instead used \emph{ternary labelings} at a point (cf.~\cite[\S 2.4]{GLMW2022}) to encapsulate information similar to that contained in the codeword of a region, albeit in a more general setting.

For any system of affine-linear equations, each equation of the form (\ref{eqn:affine-linear}) for some non-zero vector ${\bf w}$ and some $b\in\bb R$, the solution set of the system is either empty or agrees with an \emph{affine subspace} of $\bb R^n$. Hence, the intersection of some set of hyperplanes in $\bb R^n$ is either empty or an affine subspace of some dimension. We will be interested in hyperplane arrangements where the intersection of any number of the hyperplanes has a ``typical'' dimension.

\begin{defn}
\label{defn:generic}
    Let $\mathcal A =\{H_1, H_2, \ldots, H_m\}$ be a hyperplane arrangement in $\R^n$. Then $\mathcal A$ is said to be \textbf{generic} if for each of its subsets 
    \{$H_{i_1},H_{i_2}, \dots, H_{i_q} \}$, of any size $q \in \{1, \dots, m\}$, the intersection $H_{i_1} \cap H_{i_2} \cap \dots \cap H_{i_q}$ is an affine subspace with dimension $n-q$ if $q \leq n$, and is the empty set if $q>n$.
\end{defn} 

It is clear from definitions that any sub-arrangement $\mathcal A' \subset\mathcal A$ is generic if $\mathcal A$ is generic. 

\begin{remark}
\label{remark:generic-almosteverywhere}
   If $\mathcal A$ consists of $m$ hyperplanes, suppose that $H_i\in\mathcal A$ is the solution set to an equation ${\bf w}_i\cdot{\bf x} + b_i = 0$ for $1\le i\le m$. Let $W$ be the $m\times n$ matrix with the $i^{th}$ row equal to ${\bf w}_i$ and let ${\bf b}\in\bb R^m$ have the $i^{th}$ component equal to $b_i$. Identifying such pairs $(W, {\bf b})$ with points in $\bb R^{mn+m}$, it is well-known that the hyperplane arrangement so determined is generic for a full-measure subset of $\bb R^{mn+m}$ (cf.~\cite[Lemma 2.6]{GrigsbyLindsey2022}). 
\end{remark}

On occasion, given a generic hyperplane arrangement $\mathcal A$ and some $H\in\mathcal A$, it will be convenient to consider a certain \emph{induced arrangement} in $H$, obtained through intersections.

\begin{defn}
\label{defn:induced-arrangement}
    Let $\mathcal A = \{H_1,H_2,\ldots, H_m\}$ be a generic hyperplane arrangement in $\bb R^n$ and consider $H_i\in \mathcal A$ for some $1\le i\le m$. Define the ordered collection of affine subspaces $\mathcal A_{H_i} = \{H_i\cap H_1, \ldots, H_i\cap H_{i-1}, H_i\cap H_{i+1},\ldots, H_i\cap H_m\}$. We call $\mathcal A_{H_i}$ the \textbf{induced arrangement} (from $\mathcal A$) in $H_i$.
\end{defn}

By the assumption in Definition \ref{defn:induced-arrangement} that $\mathcal A$ is generic, each element of $\mathcal A_{H_i}$ is an affine subspace with dimension $n-2$. Moreover, it is possible to identify $\mathcal A_{H_i}$ with a generic hyperplane arrangement in $\bb R^{n-1}$ in a way that preserves face relations.

\subsection{Polyhedral sets} Given a subset $S\subset\bb R^n$, we use the common notation $\overline{S}$ for the closure of the set $S$ in the following definition and throughout this paper.
 
\begin{defn}
\label{defn:simplex}
A subset $P\subset\bb R^n$ is called a \textbf{\emph{polyhedral set}} in $\R^n$ if there is some finite collection of cooriented hyperplanes $\{H_1, H_2, \dots, H_k\}$ so that $P$ is the intersection of the closed positive half-spaces {--} i.e., $P = \overline{H}_1^+\cap\overline{H}_2^+\cap\ldots\cap\overline{H}_k^+$. When the polyhedral set is a bounded set, we call it a \textbf{\emph{convex polytope}}. For any subset $S\subset \R^n$, the \textbf{\emph{convex hull}} of $S$ is the intersection of all convex sets in $\R^n$ containing $S$. 

An $\textbf{n-dimensional simplex}$ is a convex polytope in $\mathbb{R}^{n} $ which agrees with the  convex hull of $(n+1)$ points, with the property that no hyperplane in $\bb R^n$ contains all $(n+1)$ points.

\end{defn}

For any region $P$ in $\bb R^n$, we say that a hyperplane $H$ \textbf{\emph{cuts through}} $P$ if there exist $p_1, p_2\in P$ such that $p_1\in H^+$ and $p_2\in H^-$. Given a polyhedral set $P$, a hyperplane $H$ is called a \textbf{\emph{supporting hyperplane}} for $P$ if $H$ does not cut through $P$ and $H\cap P\neq \emptyset$. 

\begin{defn}
\label{defn:face} 
Let $P$ be a polyhedral set. A subset $F\subset P$ is called a \textbf{\emph{face}} of $P$ if $F=\emptyset$, or $F=P$, or for some supporting hyperplane $H$ of $P$, $F=H\cap P$. 
\end{defn}

Note that Definition \ref{defn:simplex} allows for the possibility of a polyhedral set $P \subset \bb R^n$ which is contained in an affine subspace with dimension less than $n$. In that case, any hyperplane $H$ which contains $P$ as a subset will be a supporting hyperplane. We call a face of $P$ \textbf{\emph{proper}} if it is neither empty nor equal to $P$.  

The \textbf{\emph{dimension}} of a polyhedral set $P$ is the dimension of its \emph{affine hull} {--} that is, the minimal dimension of an affine subspace of $\bb R^n$ containing $P$. A face $F$ of a polyhedral set $P$ is itself a polyhedral set (e.g., if $F=H\cap P$ where $H$ is a supporting hyperplane for $P$, then give $H$ a coorientation so that $P\cap H^+ = \emptyset$, and append $H$ to a set of cooriented hyperplanes that give $P$ as a polyhedral set). If $F\ne\emptyset$, we call $F$ a $k$-dimensional face of $P$ (or, simply, a \textbf{\emph{$k$-face}}) if $F$ has dimension $k$ as a polyhedral set. A \textbf{\emph{vertex}} is a $0$-face of $P$ and a \textbf{\emph{facet}} of $P$ is a face that has dimension $\dim(P)-1$.

Every polyhedral set has an \emph{irredundant realization} as an intersection of closed half-spaces; i.e., for a polyhedral set $P\subset \bb R^n$, there are cooriented hyperplanes $H_1,H_2,\ldots,H_m$ in $\bb R^n$ so that $\displaystyle{P = \bigcap_{1\le j\le m}\overline{H}_j^+}$ and also, for every $i=1,\ldots,m$, we have 
\[P \ne \bigcap_{1\le j\le m, j\ne i}\overline{H}_j^+.\]

Let $H_1,H_2,\ldots, H_m$ provide an irredundant realization of polyhedral set $P\subset\bb R^n$, and say that the dimension of $P$ is $n$. Then $F$ is a facet of $P$ if and only if $F$ is a proper face and $F = P\cap H_i$ for some $i$, $1\le i\le m$ \cite[\S 26]{Grunbaum}. Further, it can be shown that for each $(n-k)$-face $F$ of $P$, there exist $i_1,\ldots,i_k \in\{1,2,\ldots,m\}$ so that $F = P\cap H_{i_1}\cap\ldots\cap H_{i_k}$.

For any hyperplane arrangement $\mathcal A$ in $\bb R^n$, the closure of any region of $\mathcal A$ is a polyhedral set of dimension $n$. As we will see below, for a generic, cooriented hyperplane arrangement $\textbf{A}$ in $\bb R^n$ with $n+1$ hyperplanes, only one region of $\textbf{A}$ has closure that is a convex polytope, and it is a simplex.

\subsection{Neural networks, fully connected layers, and stably unactivated neurons.}\label{subsec:notation}

This subsection introduces the notation and concepts related to neural networks that we use throughout the paper. We start by recalling the definition of the Rectified Linear Unit function, denoted by $\text{ReLU}$:
\[
\text{ReLU}(x)= max\{0,x\}, \text{ for any } x\in\R.
\]
For any $n\in\mathbb{N}$, we denote by  $\sigma:\R^n\to\R^n$ the function 
 that applies the ReLU function to each coordinate. That is, 
     \[
     \sigma({\bf x}):=(\max\{x_1,0\},\max\{x_2,0\}, \ldots, \max\{x_n,0\} ),
     \]
      where  ${\bf x}\in\R^n$.
Throughout the manuscript, we focus exclusively on feedforward ReLU neural networks that have fully-connected layers.

\begin{defn}
\label{defn::neural net and network-function} 
   Let $n_0, L\in\mathbb{N}$. A \textbf{feedforward ReLU neural network} $\mathcal N$ defined on $\R^{n_0}$, with $L$ \emph{layers}, is a list of $L$ positive integers $n_1,n_2,\ldots,n_L\in\bb N$, along with an ordered sequence of affine maps $A_1, A_2, \ldots, A_L$,  such that
     \[
     A_\ell: \R^{n_{\ell-1}}\to \R^{n_\ell}
     \]
     for each $\ell =1, \ldots, L$. Each feedforward ReLU neural network produces a function, which we call the associated \textbf{network function} $F\colon\mathbb{R}^{n_{0}}\to\mathbb{R}^{n_{L}}$, defined by
\begin{equation}\label{eq: network function}
 F({\bf x}):= A_{L}\circ\sigma\circ A_{L-1}\circ\ldots\circ\sigma\circ A_{1}(\bf x),    
\end{equation}
with $\sigma$ as above, applying ReLU coordinate-wise. We commonly abbreviate, calling $\mathcal N$ a \textbf{ReLU neural network} (or, simply, ReLU network), and $(n_0, n_1, \ldots, n_L)$ is called the \textbf{architecture} of $\mathcal N$. 
For each $\ell=1, 2, \ldots, L$, the $\ell$-th \textbf{layer map} $F_\ell: \R^{\ell-1}\to\R^{\ell}$ of the ReLU neural network defined above is given by 
\[
F_\ell:=\begin{cases}
  \sigma\circ A_{\ell}, &\text{ for } \ell =1, \ldots, L-1; \\
  A_L, &\text{ for } \ell =L.
\end{cases}
\]

\end{defn}

Given a ReLU network with architecture $(n_0,n_1,\ldots,n_L)$, a \textbf{\emph{neuron}} of the network (or, \emph{neuron in layer $\ell$}) is a pair of indices $(\ell, j)$ with $1\le \ell\le L$ and $1\le j\le n_\ell$. Hence, the coordinate function of $F_\ell$ \emph{corresponding to a neuron}, $(\ell, j)$, refers to composition of projection to coordinate $j$ in $\bb R^{n_\ell}$ with $F_\ell$.

As our neural networks have ReLU activation functions, for each $\ell=1,2,\ldots, L-1$ the image of the layer map $F_\ell$ is contained in the closed positive orthant \[\bb R^{n_\ell}_{\ge 0} = \{(x_1,\ldots,x_{n_\ell})\in\bb R^{n_\ell}\ |\ x_i\ge0,\ 1\le i\le n_\ell\}.\] Furthermore, for any $n$, $\bb R^{n}_{\ge0}$ may be viewed as a polyhedral set, determined by those hyperplanes that are defined by a coordinate being zero (the positive half-space being where that coordinate is positive). In this manuscript, our discussion will involve faces of $\bb R^n_{\ge0}$. Given $I\subset\{1,2,\ldots,n\}$, define $\mathtt{X}_{I} = \{(x_1,x_2,\ldots,x_n)\in\bb R^{n}\ |\ x_{j}=0 \text{ for each } j\not\in I\}$, and define $\halfax{I} = \mathtt{X}_{I}\cap\bb R^n_{\ge0}$. The $k$-faces of $\bb R^n_{\ge0}$ are described by those $\halfax{I}$ such that $|I| = k$. In the case that $I = \{i\}$, for some $1\le i\le n$, we call $\halfax{\{i\}}$ a (coordinate) half-axis and will write simply $\halfax{i}$.

\textbf{Fully-connected layers.} The \emph{hidden layers} of a ReLU neural network refers to all layers $\ell$ with $0<\ell< L$. For a ReLU neural network with architecture $(n_0,n_1,\ldots, n_L)$, consider an $n_{\ell}\times n_{\ell - 1}$ matrix $W_{\ell}$, for each $\ell=1,2,\ldots,L$. We say the matrix $W_{\ell}$ consists of the \emph{weights} in layer $\ell$, and those in some row $j$ are the weights corresponding to neuron $(\ell,j)$. In addition, consider a vector ${\bf b_{\ell}}\in\R^{n_\ell}$ for each $\ell$, the \emph{bias vector} in layer $\ell$. We say a neural network with architecture $(n_0,n_1,\ldots, n_L)$ has $L$ \textbf{fully-connected layers}  if, for each $\ell =1, 2, \ldots, L$,  the affine function $A_{\ell}\colon \mathbb{R}^{n_{\ell - 1}}\to\mathbb{R}^{n_{\ell}}$ is given by
    \[
    A_{\ell}({\bf x})=W_{\ell}\cdot{\bf x}+{\bf b_{\ell}},
    \]
     for any ${\bf{x}}\in\mathbb{R}^{n_{\ell-1}}$. For each $\ell =1, \ldots, L$, we denote by $W_{\ell,j}$ the $j^{th}$ row of the weight matrix $W_\ell$ and  by ${ b}_{\ell,j}$ the  $j^{th}$ component of the bias vector ${\bf b}_{\ell}$, $j=1, \ldots, n_{\ell}$.  
     
 A feedforward neural network with fully-connected layers is parameterized by a parameter space whose dimension equal to the total number of weights and biases in all layers of that neural network.

 \begin{defn}
     Define the Euclidean space $\Omega:=\R^{D}$ to be the \textbf{parameter space} for neural networks of architecture $(n_0, n_1, \ldots, n_L)$,  where a parameter $\theta\in\Omega$ is given by the set of all weight matrices and bias vectors, i.e.
\[
\theta:=(W_1,{\bf b}_1, W_2, {\bf b}_{2}, \ldots, W_L, {\bf b}_L)
\]
and 
\[
D:=\sum_{i=1}^{L}(n_{i-1}+1)n_i
\]
 is the dimension of the parameter space.
 \end{defn}   

\begin{remark}
For each $\theta$ in $\Omega=\R^D$, we denote by $F^\theta$ the network function associated with the ReLU neural network with $L$ fully-connected layers, determined by the weights and biases in $\theta$. Note that for each $\theta\in\Omega$, $F^\theta$ is a piecewise linear function. The superscript $\theta$ will be dropped when clear from the context.
\end{remark}

 \begin{defn}[]
 \label{defn: A1(theta)}Throughout this paper, we mainly focus on the interaction between the first and the second hidden layers.  For a neural network with architecture $(n_0, n_1, \ldots, n_L)$, if $\theta\in\R^D$ is understood, we denote by $\emph{\textbf{A}}_1(\theta)$ the cooriented hyperplane arrangement in $\R^{n_0}$ determined by the components of $\theta$ from the first layer map $F_1$.  For example, when $\theta =(W_1,{\bf b}_1, W_2, {\bf b}_{2}, \ldots, W_L, {\bf b}_L)$  then $\emph{\textbf{A}}_1(\theta)=\{H_1, H_2, \ldots, H_{n_1}\}$, where 
\[
H_j=\{{\bf x}\in\R^{n_0}|W_{1,j}\cdot {\bf x}+{ b}_{1,j}=0\}
\]
for each $1\leq j\leq n_1$, and the coorientation on $H_j$ is the natural one determined by $W_{1,j}$ and $b_{1,j}$, as indicated above Definition \ref{defn:coorientation}.
 \end{defn}

\begin{defn} 
\label{defn:stably-unactivated}
Let $\theta\in\Omega=\R^D.$ We say a neuron in the $\ell$-th layer is \textbf{stably unactived} at the parameter $\theta$ if there exists an open set $O\subset\Omega$ containing $\theta$ such that for every $u\in O$,
for all ${\bf x}\in\R^{n_0}$ the component of the output corresponding to that neuron in $F^u_\ell\circ \ldots\circ F^u_2\circ F^u_1({\bf x})$ is zero. 
\end{defn}

Given a neuron in layer $\ell$, for $1<\ell<L$, let $H$ denote the hyperplane associated with that neuron when the neural network has parameter $\theta$, with the natural coorientation determined by $\theta$ as in Definition \ref{defn: A1(theta)}. If the given neuron is stably unactivated at $\theta$ then $\operatorname{Im}(F^\theta_{\ell-1}\circ\ldots\circ F^\theta_2\circ F^\theta_1) \subset H^{-}$, as being in a negative half-space is an open condition. In fact, the neuron is stably unactivated at $\theta$ if and only if $\operatorname{Im}(F^\theta_{\ell-1}\circ\ldots\circ F^\theta_2\circ F^\theta_1) \subset H^{-}$, since distance between the image and hyperplane, two closed subsets, is continuous in the parameter.

\section{Supporting lemmas on hyperplanes}\label{sec:lemmas}

The first few lemmas below state classical results about combinatorial questions on hyperplane arrangements and their regions, and provide some basic observations about the code of a cooriented hyperplane arrangement.\footnote{Recall that if $\textbf{A}$ is a cooriented arrangement of $m$ hyperplanes then $\operatorname{code}(\textbf{A})$ is the subset of $\{+,-\}^m$ such that $(c_1,c_2,\ldots,c_m)\in\operatorname{code}(\textbf{A})$ precisely when there exists $R\ne\emptyset$, a region of $\textbf{A}$, so that $R\subset H_1^{c_1}\cap H_2^{c_2}\cap \ldots\cap H_m^{c_m}$.} Near the end of the section we discuss a partition of certain collections of hyperplanes that aids our discussion of stably unactivated neurons.

For the remainder of the paper, given a hyperplane arrangement $\mathcal A$ in $\bb R^n$ we use $r(\mathcal A)$ to denote the number of regions of $\mathcal A$, and $b(\mathcal A)$ to denote the number of bounded regions (i.e., those with compact closure). Lemmas \ref{lem:not-more-hyperplanes-than-n} and \ref{lem:more-hyperplanes-than-n}, which we write separately for emphasis, are well-known and follow from a theorem of Zaslavsky \cite{Zaslavsky} (cf.~ \cite[Section 2]{Stan07}).

\begin{lem}
    \label{lem:not-more-hyperplanes-than-n}
    Let $\mathcal A = \{H_1, H_2, \ldots, H_m\}$ be a generic hyperplane arrangement in $\bb R^n$, with $m\le n$. Then $r(\mathcal A) = 2^m$ and $b(\mathcal A) = 0$.
\end{lem}
\begin{lem}
    \label{lem:more-hyperplanes-than-n}
    Let $\mathcal A = \{H_1, H_2, \ldots, H_m\}$ be a generic hyperplane arrangement in $\bb R^n$, with $m \ge n$. Then \[r(\mathcal A) = \sum_{k=0}^n\binom{m}{k} \quad\text{and}\quad b(\mathcal A) = \binom{m-1}{n}.\]
\end{lem}

As an illustration of Lemma \ref{lem:more-hyperplanes-than-n}, refer to Figure \ref{fig:facet sharing}, which shows a generic arrangement of 4 hyperplanes in $\bb R^2$. The number of regions of that arrangement is $\binom{4}{0}+\binom{4}{1}+\binom{4}{2} = 11$, and the number of bounded regions is $\binom{3}{2} = 3$.

\begin{cor}
\label{cor:unique-bounded-region}
    The number of regions of a generic hyperplane arrangement of $n+1$ hyperplanes in $\bb R^n$ is $2^{n+1}-1$. Exactly one of these regions is bounded.
\end{cor}

\begin{lem}
\label{lem:nPlus1-hyperplanes}
    Suppose that $\textbf{A}$ is a generic, cooriented hyperplane arrangement of $n+1$ hyperplanes in $\bb R^n$. Then the only element from $\{+,-\}^{n+1}$ that is not contained in $\operatorname{code}(\textbf{A})$ is the negation of the codeword representing the unique bounded region.
\end{lem} 
\begin{proof}
    Given $i$ with $1\le i\le n+1$, let $\textbf{A}_i$ be the cooriented subarrangement of $\textbf{A}$ obtained by excluding the $i^{th}$ hyperplane $H_i$ (and leaving the relative ordering of the other hyperplanes unchanged). The intersection of the $n$ hyperplanes in $\textbf{A}_i$ is a single point $v_i$ (one of the vertices of the bounded region of $\textbf{A}$, a simplex, guaranteed by Corollary \ref{cor:unique-bounded-region}), and we have that $v_i$ must be in $H_i^{\epsilon_i}$ for some $\epsilon_i\in\{+,-\}$. Note that $\epsilon_i$ is determined for every $1\le i\le n+1$. Moreover, $(\epsilon_1,\epsilon_2,\ldots,\epsilon_{n+1})$ must be the codeword of the simplex of $\textbf{A}$. 

    Since $r(\textbf{A}_i) = 2^n$, we have that $\operatorname{code}(\textbf{A}_i) = \{+,-\}^n$. Furthermore, there is a neighborhood $V$ of $v_i$ that does not intersect $H_i$. Say that a region $R$ of $\textbf{A}_i$ has codeword $(c_1,c_2,\ldots,c_n)$ in $\operatorname{code}(\textbf{A}_i)$. Then, the region of $\textbf{A}$ that contains the points in $R\cap V$ will have codeword $(c_1,\ldots,c_{i-1},\epsilon_i,c_i,\ldots,c_n)$. 
    
    Since every element of $\{+,-\}^n$ is in $\operatorname{code}(\textbf{A}_i)$, every element of $\{+,-\}^{n+1}$ which has $\epsilon_i$ in position $i$ is contained in $\operatorname{code}(\textbf{A})$, and this holds for all $1\le i\le n+1$. The only ${\bf c}\in\{+,-\}^{n+1}$ which does not satisfy this for some $i$ is $(-\epsilon_1,-\epsilon_2,\ldots,-\epsilon_{n+1})$, which means that we have proven that there are $2^{n+1}-1$ codewords in $\operatorname{code}(\textbf{A})$, excluding $(-\epsilon_1,-\epsilon_2,\ldots,-\epsilon_{n+1})$. Since $r(\textbf{A}) = 2^{n+1}-1$, there cannot be a region with codeword $(-\epsilon_1,-\epsilon_2,\ldots,-\epsilon_{n+1})$.
\end{proof}
\begin{cor}
\label{cor:codeword-simplex}
    Let $R$ be a region of a generic, cooriented hyperplane arrangement $\textbf{A}$ consisting of $n+1$ hyperplanes in $\bb R^n$. Suppose that $\operatorname{code}(R) = (+,+,\ldots,+)$ and that the intersection of $\overline R$ with the unique simplex of $\textbf{A}$ is a $k$-dimensional face for some $0 \le k \le n$. Then the number of negative signs in the codeword of the bounded region is $n-k$; if $0< k < n-1$, then every codeword with exactly one positive sign is in the code of $\textbf{A}$ and represents an unbounded region of $\textbf{A}$.
\end{cor}

\begin{proof}
    Let $B$ be the unique bounded region of $\textbf{A}$. Each $k$-dimensional face of $B$ is contained in the intersection of $n-k$ hyperplanes in $\textbf{A}$. In order to pass through facets from $R$ to $B$, one must pass through each of these hyperplanes once. This changes the codeword in $n-k$ positions, resulting in $\operatorname{code}(B)$ having $n-k$ negative signs.

    Now, if $0 < k < n-1$ then $1 < n-k < n$. By Lemma \ref{lem:nPlus1-hyperplanes}, every element of $\{+,-\}^{n+1}$ is in $\operatorname{code}(\textbf{A})$ except for $-\operatorname{code}(B)$. Since $\operatorname{code}(B)$ has $n-k\ge 2$ negative signs, $-\operatorname{code}(B)$ cannot have exactly one positive sign. Finally, $B$ is the only bounded region and $\operatorname{code}(B)$ has $k+1\ge 2$ positive signs. Hence, every word with exactly one positive sign represents some unbounded region of $\textbf{A}$.
\end{proof}

We briefly highlight the last conclusion in Corollary \ref{cor:codeword-simplex}, as it helps in the proof of the main theorem. For any $1\le i\le n+1$, let ${\bf c}_i\in\{+,-\}^{n+1}$ be such that ${\bf c}_i$ is positive in its $i^{th}$ component and negative in all other components. Under the assumptions in Corollary \ref{cor:codeword-simplex}, if $0 < k < n-1$, then there is an unbounded region $R_i$ of $\textbf{A}$ so that $\operatorname{code}(R_i) = {\bf c}_i$. In addition, if $k=0$ or $k=n-1$, then this statement is true for all but one value of $i$ in $\{1,2,\ldots, n+1\}$.

\begin{lem}
\label{lem:more-hyperplanes}
    Suppose that $\textbf{A}$ is a generic, cooriented hyperplane arrangement of $m$ hyperplanes in $\bb R^n$ with $m > n$. Let $R$ be one of the bounded regions of $\textbf{A}$. Then $-\operatorname{code}(R)$ is not in the code of $\textbf{A}$.
\end{lem}
\begin{proof}
    Suppose that $\textbf{A} = \{H_1,H_2,\ldots, H_m\}$ and let $\operatorname{code}(R) = (c_1,c_2,\ldots,c_m)$.  There are at least $n+1$ facets for the closure $\overline{R}$, and so at least $n+1$ hyperplanes of $\textbf{A}$ that contain a facet of $\overline{R}$. We claim that there is a choice of exactly $n+1$ of these hyperplanes, giving a generic subarrangement $\textbf{A}' = \{H_{i_1},\ldots,H_{i_{n+1}}\} \subset \textbf{A}$, such that $R$ is a subset of the unique bounded region of $\textbf{A}'$ guaranteed by Corollary \ref{cor:unique-bounded-region}. Call this bounded region $R'$. Using the coorientation and ordering on $\textbf{A}'$ given by that on $\textbf{A}$, we have that $\operatorname{code}(R') = (c_{i_1},\ldots,c_{i_{n+1}})$. Every region of $\textbf{A}$ is formed through a subdivision of a region of $\textbf{A}'$ with some of the hyperplanes not in $\textbf{A}'$ (potentially a \emph{trivial} subdivision if no hyperplane in $\textbf{A}\setminus\textbf{A}'$ cuts through the region). Hence, components $i_1,i_2,\ldots$, $i_{n+1}$ of the codeword for a region of $\textbf{A}$ agree with the codeword of a region of $\textbf{A}'$. By Lemma \ref{lem:nPlus1-hyperplanes}, we know that $-\operatorname{code}(R') \not\in\operatorname{code}(\textbf{A}')$ and so $-\operatorname{code}(R)$ cannot be in $\operatorname{code}(\textbf{A})$.
    
    In order to prove that the claimed subarrangement $\textbf{A}'$ exists, first note that it is immediate if $m = n+1$ (and so $\textbf{A}' = \textbf{A}$), by Lemma \ref{lem:nPlus1-hyperplanes}. Moreover, the claim holds when $n=1$, since every closed bounded region has exactly $2$ facets in that case. Now, suppose that $n\ge2$ and that $m > n+1$. Let $k$ be the number of facets of $\overline{R}$ and let $\{H_{i_1}, H_{i_2},\ldots, H_{i_k}\}\subset\textbf{A}$ denote those hyperplanes that contain these $k$ facets. We may assume that $k>n+1$, since we take $\textbf{A}'=\{H_{i_1}, H_{i_2},\ldots, H_{i_k}\}$ and $R'=R$ otherwise. We will proceed by induction on both $m$ and the dimension $n$.
    
    For each $1\le j\le k$, exactly one region $S_j$ of $\textbf{A}$ has the property that $\overline{R}\cap\overline{S}_j$ is contained in $H_{i_j}$ and is a facet of $\overline{R}$. If $S_1$ is bounded then define ${\bf A}_1 = \textbf{A} \setminus \{H_{i_1}\}$ and let $R_1$ be the interior of $\overline{R}\cup \overline{S}_1$. As ${\bf A}_1$ consists of $m-1$ hyperplanes and $R_1$ is a bounded region of $\textbf{A}_1$ containing $R$, the result follows by induction. Hence, we assume that $S_1$ is unbounded. %

    Recalling Definition \ref{defn:induced-arrangement}, consider $\textbf{A}_{H_{i_1}}$, the induced arrangement in $H_{i_1}$ from $\textbf{A}$. Using that $\textbf{A}$ is generic, we may identify $\textbf{A}_{H_{i_1}}$ with a hyperplane arrangement in $\bb R^{n-1}$, the elements $\{H_{i_1}\cap H_\ell\ |\ \ell\ne i_1, 1\le \ell\le m\}$ being identified with hyperplanes in that dimension. Write $T_1 = \overline{R}\cap\overline{S}_1$ for the facet of $\overline{R}$ that is contained in $H_{i_1}$ and note that $T_1$ agrees with a bounded region of $\textbf{A}_{H_{i_1}}$. By induction on dimension, there is a subarrangement of $n$ hyperplanes in $\textbf{A}_{H_{i_1}}$ whose unique bounded region contains $T_1$ as a subset. That is, we have a set of indices $\ell_1, \ldots, \ell_n$, with $i_1\ne \ell_j$ for all $1\le j\le n$, so that $T_1$ is a subset of the unique bounded region of $\{H_{i_1}\cap H_{\ell_1}, \ldots, H_{i_1}\cap H_{\ell_n}\}$. Define $\textbf{A}' = \{H_{i_1}, H_{\ell_1}, \ldots, H_{\ell_n}\}$. 

    Let $S'_1$ be the region of $\textbf{A}'$ which contains $S_1$ and has a codeword (in $\operatorname{code}(\textbf{A}')$) that agrees with components $i_1,\ell_1,\ldots,\ell_n$ of $\operatorname{code}(S_1)$. Since $S_1$ is unbounded, $S'_1$ must also be unbounded {--} we obtain it from $S_1$ by removing hyperplanes that are not in $\textbf{A}'$. Let $c_1$ be component $i_1$ of $\operatorname{code}(S_1)$, and so the region $S'_1$ is in the half-space $H_{i_1}^{c_1}$. Since $\textbf{A}'$ is a generic arrangement of $n+1$ hyperplanes and $S'_1$ is an unbounded region, on the $c_1$-side of $H_{i_1}$, that has $n+1$ facets ($n$ of them coming from those hyperplanes that intersect $H_{i_1}$ to give the $n$ facets of the bounded region we found, containing $T_1$), the region we get by crossing the facet that is in $H_{i_1}$ {--} i.e., switching only the sign of $c_1$ in the codeword {--} must be the unique bounded region of $\textbf{A}'$. By the way that we obtained the facet we crossed, a superset of $T_1$, our unique bounded region contains $R$.  
\end{proof}

\subsection{Partitioning a collection of hyperplanes}
\label{subsec:hyperplane-partition}
Let $H$ be a hyperplane in $\bb R^n$, determined by a set of continuous random variables $w_1,w_2,\ldots,w_n,b$ where, using ${\bf w}$ to denote $(w_1,w_2,\ldots,w_n)$, we have $H = \{{\bf x}\in\bb R^n\ |\ \textbf{w}\cdot{\bf x} + b = 0\}$. 

Almost surely, $w_i\ne0$ for all $i=1,2,\ldots,n$ and $b\ne0$, in which case $H$ intersects all of the coordinate axes away from the origin.  Conversely, the set of such intersection points determines $H$. If the $n$ intersection points are ${\bf q}_1, \ldots, {\bf q}_n$ then, more simply, we may refer to their one nonzero coordinate: writing $(q_1, \ldots, q_n)$ such that ${\bf q}_1 = (q_1,0,\ldots,0)$, and so on. In the setting that it is well-defined, we call $(q_1,q_2,\ldots,q_n)$ the {\bf \emph{intercept tuple}} of $H$.

Given a fixed $n$-tuple, ${\bf p} = (p_1, \ldots, p_n)$ in $\bb R^n$, such that no $p_i$ is $0$, define $\Hp{\bf p}$ to be the set of hyperplanes with a well-defined intercept tuple $(q_1, q_2,\ldots,q_n)$, such that $q_i\ne 0$ for all $1\le i\le n$ and $q_i$ is positive if and only if $p_i$ is positive. 

We separate hyperplanes in $\Hp{\bf p}$ into subsets $\mathcal P, \Su_1, \Su_2,\ldots, \Su_n$ as follows. Let $H\in \Hp{\bf p}$ and let $(q_1,\ldots,q_n)$ be its intercept tuple. For each $j=1,\ldots,n$, there is a scalar $\lambda_j > 0$ such that $q_j = \lambda_jp_j$. Define $\mathcal P\subset\Hp{\bf p}$ by 
    \[\mathcal P = \{H\in\Hp{\bf p}\ |\ \text{there is } j\ne j',\ 1\le j, j'\le n, \text{ such that } \lambda_j = \lambda_{j'}\}.\]
Also, for each $j=1,\ldots,n$, define 
    \[\Su_j = \{H\in\Hp{\bf p}\setminus\mathcal P\ |\ \lambda_j > \lambda_i\ \text{for all}\ i\ne j, 1\le i\le n \}.\]
Given $j$ with $1\le j\le n$, note that if $H\in\Su_j$ then for each $i\ne j$ we have $|q_i|=|\lambda_ip_i| < |\lambda_jp_i|$. Thus, on the $i^{th}$ coordinate axis, $i\ne j$, $H$ intercects the axis closer to the origin than the hyperplane associated to $(\lambda_jp_1,\lambda_jp_2,\ldots,\lambda_jp_n)$.

The proof of the following lemma is straightforward and we leave its verification to the reader.

\begin{lem}
\label{lem: partition} 
For any ${\bf p}\in\bb R^n$, with $p_i\ne 0$ for all $1\le i\le n$, $\Hp{\bf p}$ is partitioned by $\{\mathcal P, \Su_1, \ldots, \Su_n\}$.
\end{lem}

\begin{remark}
    For a fixed ${\bf p}$, a hyperplane $H\in\Hp{\bf p}$ is determined by $\lambda_1,\lambda_2,\ldots, \lambda_n$. 
\end{remark}

\subsection{Network layer maps and stably unactivated neurons}\label{sec:node-layer2} 

In the remainder of the section, we consider a ReLU neural network with parameter $\theta$. As in Definition \ref{defn: A1(theta)}, $W_1$ (resp.\ $\textbf{b}_1$) is the weight matrix (resp.\ bias vector) in the first layer of $\mathcal N(\theta)$. As in Definition \ref{defn::neural net and network-function}, we have the layer map $F_1:\bb R^{n_0}\to\bb R^{n_1}$ given by $F_1(\textbf{x}) = \sigma(W_1\cdot{\bf x} + {\bf b}_1)$. The affine map that sends ${\bf x}\mapsto W_1\cdot{\bf x}+{\bf b}_1$ is the \emph{pre-activation map} for the first layer. Weights and biases are considered as continuous random variables. 

\begin{remark}
    \label{remark:regions-convex-image}
Given a non-empty region $R$ of $\textbf{A}_1(\theta)$, say that $I\subset\{1,2,\ldots,n_1\}$ is the set of components in $\operatorname{code}(R)$ that are positive. Then it is clear that $F_1(\overline R) \subset \halfax{I}$. Since the pre-activation map for the first layer sends every point of $R$ to just one of the orthants of $\bb R^{n_1}$, the restriction of $F_1$ to $\overline R$ is an affine map. Therefore, as $\overline R$ is convex, $F_1(\overline R)$ is also convex for any region $R$ of the hyperplane arrangement $\textbf{A}_1(\theta)$.
\end{remark}

Let $n_1 = n_0+1$. If the hyperplane arrangement $\textbf{A}_1 = \textbf{A}_1(\theta)$ is generic, it has one bounded region (Corollary \ref{cor:unique-bounded-region}), which we call $B$. For every $i = 1,2, \ldots, n_0+1$, let ${\bf v}_i$ be the vertex of $\overline{B}$ comprising the intersection of all but the $i^{th}$ hyperplane in $\textbf{A}_1$. The image of the pre-activation map is a hyperplane in $\bb R^{n_1}$, which does intersect the $i^{th}$ coordinate axis at $W_{1,i}\cdot{\bf v}_i + b_{1,i}$ for each $1\le i\le n_0+1$. Since $\textbf{A}_1$ is generic, ${\bf v}_i$ is well-defined and not in the $i^{th}$ hyperplane. Therefore, the intercept $W_{1,i}\cdot{\bf v}_i + b_{1,i}$ is defined and non-zero; it is positive if and only if $\operatorname{code}(B)$ is positive in the $i^{th}$ component. 
\begin{remark}
\label{remark:codeword-and-intercepts}
Use $P_+=P_+(W_1,{\bf b}_1)$ for the hyperplane that is the image of the pre-activation map and write ${\bf p} = (p_0,p_1,\ldots, p_{n_0+1})$ for the intersection tuple of $P_+$. By the previous paragraph, we conclude $\operatorname{code}(B) = (\operatorname{sign}(p_1), \operatorname{sign}(p_2), \ldots, \operatorname{sign}(p_{n_0+1}))$.
\end{remark}

Consider a region $R$ of the arrangement $\textbf{A}_1$ whose closure intersects $\overline{B}$ in a common $k$-face, for some $0\le k\le n_0$, and such that $\operatorname{code}(R)$ has only positive components. Since $\operatorname{code}(R)$ is positive in every component, 
the hyperplane $P_+(W_1,{\bf b}_1)$ in Remark \ref{remark:codeword-and-intercepts} agrees with the unique hyperplane containing $F_1(R)$. In the remainder of the paper, we may refer to this hyperplane simply as $P_+$ (with an understood first layer map determined by $W_1, {\bf b}_1$).

By Corollary \ref{cor:codeword-simplex} we have that $\operatorname{code}(B)$ contains exactly $n_0-k$ negative signs and so, by the above comments, we have shown the following.

\begin{lem} 
\label{lem:intercepts-signs}
Let a neural network have an architecture $(n_0,n_0+1, n_2,\ldots,n_L)$ and suppose that $\textbf{A}_1$ is generic. If $R_+$, a region of $\textbf{A}_1$, has a codeword $\operatorname{code}(R_+)=(+,+,\ldots,+)$ and shares a $k$-face with the unique bounded region of the hyperplane arrangement, then the hyperplane in $\bb R^{n_1}$ which contains $F_1(R_+)$ as a subset intersects exactly $n_0-k$ of the coordinate axes negatively. 
\end{lem}

In Lemma \ref{lem:intercepts-signs} above, if we have that $k=n_0-1$, then by Lemma \ref{lem:nPlus1-hyperplanes} there is no region with codeword $(-,\ldots,-,+,-,\ldots,-)$, where the position of $+$ corresponds to the facet that $\overline{R}_+$ shares with $\overline{B}$. Consequently, $\ImF$ and the corresponding axis in $\bb R^{n_1}$ intersect in only the origin if $\overline{R}_+\cap\overline{B}$ is a facet of the simplex.

To connect our discussion with the previous subsection, suppose that $W_1$ and ${\bf b}_1$ are the weight matrix and bias vector, respectively, of the first layer of a neural network, and that $W_1, {\bf b}_1$ are such that $\textbf{A}_1$ is a generic hyperplane arrangement. Further, let $P_+ = P_+(W_1,{\bf b}_1)$, as in Remark \ref{remark:codeword-and-intercepts}.

\begin{lem}
\label{lem:prob-mathcalP=0}
If ${\bf p}=(p_1,\ldots,p_{n_1})$ is the intercept tuple of $P_+=P_+(W_1,{\bf b}_1)$, with assumptions as above, let $\{\mathcal P, \Su_1, \ldots, \Su_{n_1}\}$ be the partition of $\Hp{\bf p}$ discussed in Lemma \ref{lem: partition}. If $n_1 = n_0+1$, and the weights and bias corresponding to a given neuron in the second layer determine a hyperplane $H$, then $\bb P(H\in\mathcal P) = 0$. 
\end{lem}

\begin{proof}
Write the weights and bias for the given neuron in the second layer, which determine $H$, as ${\bf w}=(w_1,\ldots,w_{n_0+1})$ and $b$, respectively. If $H$ has a well-defined intercept tuple $(q_1,q_2,\ldots,q_{n_0+1})$, with only nonzero intercepts, then $q_j = -b/w_j$ for every $1\le j\le n_0+1$. As discussed above Remark \ref{remark:codeword-and-intercepts}, we have $p_j = W_{1,j}\cdot{\bf v}_j+b_{1,j}$, where ${\bf v}_j$ is the vertex obtained by intersecting all but the $j^{th}$ hyperplane in $\textbf{A}_1$. As a continuous random variable, ${\bf v}_j$ is completely determined by the weights in $W_1$ and biases in ${\bf b}_1$ (excluding those from row $j$).

As a consequence, for $j\ne j'$, the variables $\lambda_j = q_j/p_j$ and $\lambda_{j'} = q_{j'}/p_{j'}$ are identically distributed random variables (provided weights within the first layer are identically distributed, as are biases in that layer).

From definitions, we have $\bb P(H\in\mathcal P) = \bb P(\exists\ j\ne j'\ \text{s.t.}\ \lambda_j = \lambda_{j'})$. The equation $\lambda_j = \lambda_{j'}$ is equivalent to $p_{j'}q_j = p_jq_{j'}$, which is a rational equation,
		\[(W_{1,j'}\cdot{\bf v}_{j'}+b_{1,j'})bw_{j'} 
		    =   (W_{1,j}\cdot{\bf v}_{j}+b_{1,j})bw_j.\]
Therefore, the solution set has positive codimension in $\bb R^D$, where $D = n_0n_1+2n_1+1$ is the number of weights and biases. This remains true when we remove the finite set of affine algebraic curves to guarantee that $\textbf{A}_1$ is generic, and those given by $\{b=0\}$, $\{w_i = 0\}$, $i=1,2,\ldots,n_0+1$, to guarantee that $H$ has a well-defined intercept tuple.
\end{proof}

\section{Proof of the Main Results}
\label{sec:proof-main}

Consider a ReLU neural network $\mathcal N$ with architecture $(n_0,n_1,\ldots,n_L)$. Given a neuron in the second layer of $\mathcal N$, use $E\subset\Omega$ to denote those $\theta$ at which the neuron is stably unactivated. Let $E^+\subset E$ denote the event that at least one of the weights and bias associated with that neuron, written $({\bf w}|b)$, is positive. Let $H$ be the cooriented hyperplane associated to $({\bf w}|b)$. As above, $F_1 = F^{\theta}_1$ is the first layer map of $\mathcal N(\theta)$. As is implied by the discussion after Definition \ref{defn:stably-unactivated}, $E=\{\theta\in\Omega | \ImF\subset H^-\}$.

In the sequel, we fix a generic, not cooriented, hyperplane arrangement $\mathcal A_1$ of $n_1$ hyperplanes in $\bb R^{n_0}$ and a subset $\Omega_1 \subset \Omega$ which has the property that for all $\theta\in\Omega_1$, $\textbf{A}_1(\theta)$ is the arrangement $\mathcal A_1$ equipped with a coorientation. We restrict to parameters $\theta$ for $\mathcal N$ which are in $\Omega_1$. 

Our strategy for proving Theorem \ref{thm:small-n1} and Theorem \ref{thm:main} is to show that $\bb P(E\ |\ \theta\in\Omega_1)$ is independent of $\mathcal A_1$ and equal to the value in the theorem statements. We may then conclude that $\bb P(E\ |\ \theta\in\Omega_1) = \bb P(E\ |\ \textbf{A}_1(\theta) \text{ is generic})$, which equals $\bb P(E)$ by Remark \ref{remark:generic-almosteverywhere}.

First, we prove Theorem \ref{thm:small-n1} on the probability of $E$ in the case that $n_0\ge n_1$. The proof starts with the following lemma.  
\begin{lem}[]\label{Lemma:ImF1}
 Let $(n_0, n_1, \ldots, n_L)$ be  the architecture of a neural network and assume that ${\bf A}_1$ is generic. 

\begin{itemize}
    \item[(a)] If $n_0\geq n_1$, then all coordinate half-axes $\halfax{j}$, $j=1,2,\ldots,n_1$, are subsets of $\ImF$.
    \item[(b)] If $n_1 = n_0+1$,  denote by $B$ the bounded region in ${\bf A}_1$. 
  If $\operatorname{code}(B)$ has exactly one $+$ or exactly one $-$ in the $i^{th}$ component, then $\halfax{j} \subset \ImF$ if $j\ne i$; otherwise, unless $\operatorname{code}(B)$ consists of only positive components, $\halfax{j} \subset \ImF$ for all $1\le j\le n_1$. 
    
\end{itemize}

\end{lem}
\begin{proof}[Proof of Lemma \ref{Lemma:ImF1}]
Let $\textbf{A}_1=\{H_1, \ldots, H_{n_1}\}$ and denote by $R$ a region of ${\bf A}_1$ which has a codeword with exactly one positive sign, in the $j^{th}$ component for some $1\le j\le n_1$. That is, $j$ is the unique index so that $R\subset H_j^+$. By Remark \ref{remark:regions-convex-image}, $F_1(\overline{R})$ is convex and a subset of $\halfax{j}$. It follows that if $F_1(\overline{R})$ is unbounded and $R$ is adjacent to a region whose codeword has only negative components (which is sent to the origin by $F_1$), then $\halfax{j} \subset \ImF$.

In the case of $n_0\geq n_1$, by Lemma \ref{lem:not-more-hyperplanes-than-n} each of the $2^{n_1}$ regions of $\textbf{A}_1$ are unbounded, so $R$ is unbounded. It is, then, straightforward to check that the coordinate function of the pre-activation map, given by ${\bf x}\mapsto W_{1, j}\cdot{\bf x} + b_{1,j}$, is unbounded along any infinite ray in $R$. Thus, $F_1(\overline{R})$ is unbounded. Since every element of $\{+,-\}^{n_1}$ is in $\operatorname{code}(\textbf{A}_1)$, there is a region with all components being negative, sent to the origin by $F_1$. As this region of $\bb R^{n_0}$ is determined by being on the negative side of all of its facets, it must be adjacent to $R$, proving that $\halfax{j}\subset\ImF$. 

Now let us show part (b) of Lemma \ref{Lemma:ImF1} when $n_1=n_0+1$.
By Corollary \ref{cor:unique-bounded-region}, there are $2^{n_0+1}-1$ regions of ${\bf A}_1$, and exactly one of these regions is bounded, denoted by $B$.

If  $\operatorname{code}(B)$ has exactly one negative sign in the $i^{th}$ component then $-\operatorname{code}(B)$, which has exactly one positive sign in the $i^{th}$ component, is not in $\operatorname{code}(\textbf{A}_1)$ by Lemma \ref{lem:nPlus1-hyperplanes}. Hence, the only point in $\halfax{i}\cap\ImF$ is the origin. If $\operatorname{code}(B)$ has exactly one positive sign in the $i^{th}$ component, then $\halfax{i}\cap\ImF$ is compact since $\overline{B}$ is compact\footnote{In fact, the intersection of $\halfax{i}\cap\ImF$ is a line segment starting from the origin in this case.}. By Lemma \ref{lem:nPlus1-hyperplanes}, $-\operatorname{code}(B)$ 
is the only word in $\{+,-\}^{n_0+1}\setminus\operatorname{code}(\textbf{A}_1)$, and so in both cases all the other codewords with exactly one positive component represent an unbounded region of ${\bf A}_1$ which shares a facet with the region that is negative in all components. As in the proof for part (a), we find that $\halfax{j}\subset\ImF$ for all $j\ne i$. 

Finally, the remaining cases to consider are when $\operatorname{code}(B)$ has $k$ negative components, with $k > 1$ and $k \ne n_0$. In these cases, by Corollary \ref{cor:codeword-simplex}, each codeword with exactly one positive sign will be in $\operatorname{code}(\textbf{A}_1)$ and will represent
an unbounded region of $\textbf{A}_1$.
Moreover, as $k\ne0$, a non-empty region whose codeword has all negative components is sent to the origin and we have that $\halfax{j}\subset\ImF$ for all $j=1,2,\ldots, n_0+1$. 
 \end{proof}
 
\begin{proof}[\textbf{Proof of Theorem \ref{thm:small-n1}}]

Let $H$ be the hyperplane, determined by $({\bf w}|b)$ and associated with our given node in layer 2. Then $\ImF\subset H^-$ requires that $b<0$ (since the origin is in $\ImF$). Furthermore,
by part (a) of Lemma \ref{Lemma:ImF1}, we know that $\halfax{j}\subset\ImF$ for all $1\le j\le n_1$.
Consequently, we must have that each intersection that $H$ has with a coordinate axis in $\bb R^{n_1}$ is negative. Thus all coordinates of ${\bf w}$ are negative, almost surely. As $\ImF\subset\bb R^{n_1}_{\ge0}$, knowing ${\bf w}<0$ and $b<0$ is also sufficient for $\ImF\subset H^{-}$. Hence,
    \[\bb P(E|\theta\in\Omega_1) = \bb P(\{{\bf w} < 0\} \cap \{b < 0\}) = \left(\frac{1}{2}\right)^{n_1+1}.\]

\end{proof}

Next, we focus on the case that $n_1=n_0+1$. We know there is a unique bounded region of $\textbf{A}_1$ according to Corollary \ref{cor:unique-bounded-region}, the closure of which is an $n_0$-dimensional simplex. As before, we use $B$ to denote the bounded region, then $\overline{B}$ is the simplex.
Based on how the closed positive half-spaces of the hyperplanes in $\textbf{A}_1$ intersect $\overline{B}$ and considering that the weights and biases are independent and symmetrically distributed around the origin, there are $2^{n_1} = 2^{n_0+1}$ possible equally likely cases for the coorientation. We call each case a {\bf \textit{configuration}}, denoted by $C_i$, $i=0,\ldots, 2^{n_0+1}-1$.

To establish our main result we require a key lemma, stated below, which records the sum of the conditional probabilities of $E^+$ given that  $\textbf{A}_1$ is  a particular configuration. Breaking down the probability $\bb P(E^+|\theta\in\Omega_1)$ according to the configuration is essential in the proof of the lemma.
\begin{lem}\label{Lemma: Sum of delta}
Let $\delta_i = \P( E^+ | \theta\in C_i)$, $i=0,\ldots, {2^{n_1}-1}$. If $n_1=n_0+1$, then we have 
\[
\sum_{i=0}^{2^{n_1}-1}\delta_i =\frac{1}{2^{n_1}}.
\]
\end{lem}
The proof of this lemma will be postponed to Section \ref{sec:configs}.
Next, we prove our main result Theorem \ref{thm:main}. 

\begin{proof}[\textbf{Proof of Theorem \ref{thm:main}}]
   Based on the definition of $C_i$, for each $i=0,\ldots, 2^{n_1}-1$, it is clear that $\{C_0,C_1, C_2, \ldots, C_{2^{n_1}-1}\}$ is a set of mutually exclusive and collectively exhaustive events.   By the total probability formula, $\P(E^+|\theta\in\Omega_1)$ is given by:
   \[
   \P(E^+|\theta\in\Omega_1)= \sum_{i=0}^{2^{n_1}-1}\P(E^+|\theta\in C_i)\P(\theta\in C_i)=\sum_{i=0}^{2^{n_1}-1}\delta_i\P(\theta\in C_i).
   \]
   Let $E^-$ denote the event that all weights and bias in $({\bf{w}}|b)$ are negative, then $E^-$ and $E^+$ are mutually exclusive and $E^{+}\cup E^{-} =E$.  
   In addition, as the weights and bias are independent and symmetric distributed around the origin, we have
   $\P(\theta\in C_i)=\frac{1}{2^{n_1}}$ for any $i=0, 1, \ldots, 2^{n_1}-1$ and $P(E^-|\theta\in\Omega_1)=\frac{1}{2^{n_1+1}}$. 
   Thus 
\begin{align*}
\P(E|\theta\in\Omega_1)&=\P(E^-|\theta\in\Omega_1)+\P(E^+|\theta\in\Omega_1)=\frac{1}{2^{n_1+1}}+\sum_{i=0}^{2^{n_1}-1}\delta_i\P(\theta\in C_i)\\
&=\frac{1}{2^{n_0+2}}+\frac{1}{2^{n_0+1}}\frac{1}{2^{n_0+1}}=\frac{2^{n_0}+1}{4^{n_0+1}}.
\qedhere
\end{align*}
\end{proof}

\hfill

\section{Proof of Lemma \ref{Lemma: Sum of delta}}\label{sec:configs}

This section is devoted to proving Lemma \ref{Lemma: Sum of delta}. Throughout the section, we consider a ReLU neural network where $n_1=n_0+1$; that is, the network has an architecture of $(n_0,n_0+1,n_2,\ldots,n_L)$ with $L\ge 2$. Otherwise, assumptions are the same as in Section \ref{sec:proof-main}.

To describe the \emph{configurations} $C_0,C_1,\ldots, C_{2^{n_1}-1}$ which were referenced in Lemma \ref{Lemma: Sum of delta}, we index the elements of $\{+,-\}^{n_0+1}$ as follows. Set ${\bf c}_0\in\{+,-\}^{n_0+1}$ to be the element in which every component is positive. Next, for each $i\in\{1,2,\ldots, n_0+1\}$, let ${\bf c}_i\in\{+,-\}^{n_1}$ be the element with $i^{th}$ component negative, and all other components positive. For each $i\in\{n_0+2, \ldots, 2n_0+2\}$, define ${\bf c}_i = -{\bf c}_{i-n_0-1}$. Finally, index the remaining elements of $\{+,-\}^{n_1}$ in an arbitrary order, ${\bf c}_{2n_0+3},\ldots, {\bf c}_{2^{n_1}-1}$.

As in the previous sections, we use $B$ to denote the
the unique bounded region of $\mathcal A_1$ and use $\overline{B}$ to denote the closure of $B$. That is, $\overline{B}$ is the simplex of  $\mathcal A_1$. Now, for $i=0,1,\ldots, 2^{n_1}-1$ define $C_i = \{\theta\in\Omega_1\ |\ \operatorname{code}(B) = {\bf c}_i\}$ where $\Omega_1$ is a subset of $\Omega=\mathbb{R}^D$ with full measure. Finally, we define the conditional probabilities $\delta_i = \bb P(E^+\ |\ \theta\in C_i)$.

\begin{figure}[t]
\begin{tikzpicture}[>=stealth, scale=0.6]
    \draw[<->,cornflower!80!blue] (-1,3) --node[at start,right]{{\footnotesize $H_1$}} (-1,-3);
    \draw[<->,cornflower!80!blue] (-3,-2.6) --node[at end,above left]{{\footnotesize $H_2$}} (3,1);
    \draw[<->,cornflower!80!blue] (-3,1.128) --node[at end,below left]{{\footnotesize $H_3$}} (3, 1.128 - 2.4);

    \node[cornflower!80!blue] at (-0.31,-0.45) {{\tiny $--+$}};
    \node[cornflower!80!blue] at (-2.0,-0.45) {{\tiny $+-+$}};
    \node[cornflower!80!blue] at (-2.0,1.4) {{\tiny $+--$}};
    \node[cornflower!80!blue] at (-2.0,-2.6) {{\tiny $+++$}};
    \node[cornflower!80!blue] at (0.61,0.95) {{\tiny $---$}};
    \node[cornflower!80!blue] at (0.61,-1.85) {{\tiny $-++$}};
    \node[cornflower!80!blue] at (2.15,-0.45) {{\tiny $-+-$}};
    
    \node[inner sep=0pt] (surface) at (8,0)
    {\includegraphics[width=0.35\textwidth]{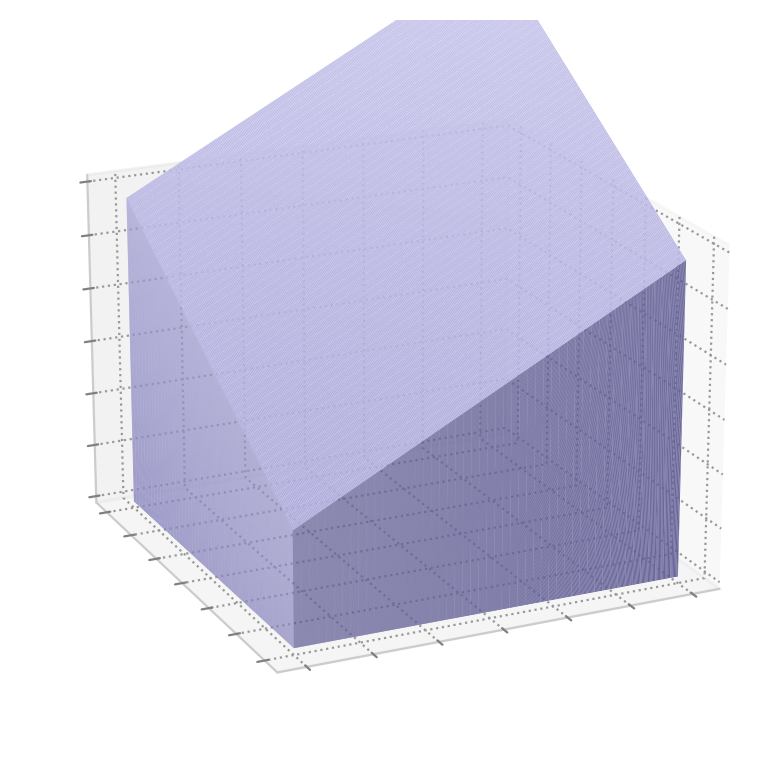}};
\end{tikzpicture}
\caption{Hyperplane arrangement $\textbf{A}_1$ (left) and image of first layer map $F_1:\bb R^2\to\bb R^3$ (right) for example $\theta\in C_6$; $n_0=2$, $n_1=3$.}
\label{fig:vertex-config}
\end{figure}
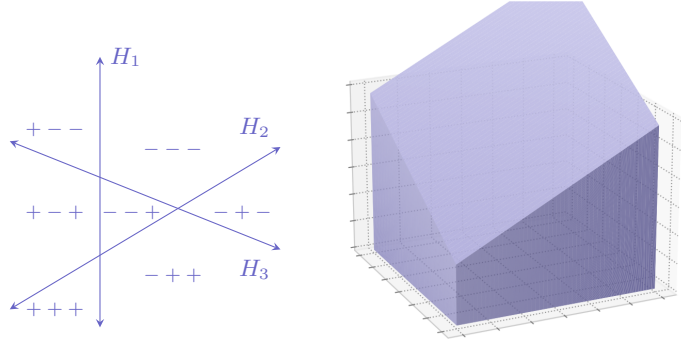

\begin{remark}
    If $\textbf{A}_1$ is such that $(+,+,\ldots,+)\in\operatorname{code}(\textbf{A}_1)$, use $R_+$ to denote the region of $\textbf{A}_1$ which has this codeword. The indexing of the configurations above is such that $\theta\in\Omega_1$ is in one of $C_1,C_2,\ldots, C_{n_0+1}$ if and only if there is a such a region $R_+$ for $\textbf{A}_1(\theta)$ and $\overline{R}_+ \cap \overline{B}$ is a facet of $\overline{B}$, as in Figure \ref{fig:facet-config}. On the other hand, $\theta$ is in one of $C_{n_0+2},C_{n_0+3},\ldots, C_{2n_0+2}$ if and only if there is a region $R_+$ of $\textbf{A}_1(\theta)$ and $\overline{R}_+ \cap \overline{B}$ is a vertex of $\overline{B}$, as in Figure \ref{fig:vertex-config}. 
\end{remark}

As discussed above, the neuron of the second layer in question is stably unactivated if and only if $\ImF \subset H^-$. In particular, it is necessary that $\ImF\cap H = \emptyset$. In addition, to ensure that $({\bf w}|b)$ are drawn from the event $E^+$, rather than $E^-$, we need (almost surely) that $H$ cuts through the positive orthant $\bb R^{n_1}_{\ge0}$, as we now explain.

\begin{remark}
\label{rem:H-positiveorthant}
  The event $E^+$ occurs when the weights and bias are not all negative. Note that, almost surely,\footnote{One of the weights, or the bias, being equal to zero happens with probability zero.} this implies that $H$ cuts through the positive orthant. To see this, if some component of ${\bf w}$, say $w_i$, and $b$ have opposite signs, then any point in $\halfax{i}$ with a sufficiently large $i^{th}$ coordinate will not be in the same half-space of $H$ as the origin, which implies that $H$ cuts through $\bb R^{n_1}_{\ge0}$. The only remaining possibility is that the weights and bias are all positive. But then the positive orthant is in $H^+$, not $H^-$, and thus $E$ cannot occur.
\end{remark}

By our indexing, we see that for each $i=2n_0+3, \ldots, 2^{n_1}-1$, if $\theta\in C_i$ then the codeword of the bounded region $B$ either consists of all negative components, or it has at least two negative and at least two positive components.  By part (b) of Lemma \ref{Lemma:ImF1}, we can conclude that the positive half-axes $\halfax{j}$, $j=1,2,\ldots,n_0+1$, are each a subset of $\ImF$. By Remark \ref{rem:H-positiveorthant}, we obtain the following result.

\begin{remark}\label{rem:zero-deltas}
For $j=2n_0+3, \ldots, 2^{n_1}-1$, we have $\delta_j=0$. Thus, the proof of Lemma \ref{Lemma: Sum of delta} is reduced to showing that 
\[
\sum_{i=0}^{2n_0+2}\delta_i=\frac{1}{2^{n_1}} \text{ where } n_1=n_0+1.
\]
\end{remark}

In the following subsections, we consider $\ImF$ when $\theta$ is in one of the configurations $C_0, C_1,\ldots, C_{2n_0+2}$ and we provide a useful reformulation of the condition that $H$ satisfies $\ImF\subset H^-$. For this purpose, the configurations are separated into three cases depending on the number of negative components of the codeword of the unique bounded region $B$. Moreover,  in each of the configurations $C_0,C_1,\ldots,C_{2n_0+2}$, there is a positive region, denoted $R_{+}$, whose codeword is $\operatorname{code}(R_{+}) = (+,+,\ldots,+)$. First, configuration $C_0$ is considered on its own, in which case $R_+ = B$. Secondly, we consider $C_1,C_2,\ldots,C_{n_0+1}$, in which cases $\overline{B}\cap \overline{R}_{+}$ is a facet of $\overline{B}$. Finally, we consider $C_{n_0+2},\ldots, C_{2n_0+2}$, where $\overline{B}\cap \overline{R}_{+}$ is a vertex of  $\overline{B}_{+}$. Our reformulation in the second case allows us to compute the sum $\sum_{i=1}^{n_0+1}\delta_i$. By the first and third cases, we are able to relate these $\delta_i$'s to the sum of the others and arrive at our conclusion. After considering these three cases, we combine all the findings at the end of Section \ref{subsec:vertex-case} to conclude the proof of Lemma \ref{Lemma: Sum of delta}.

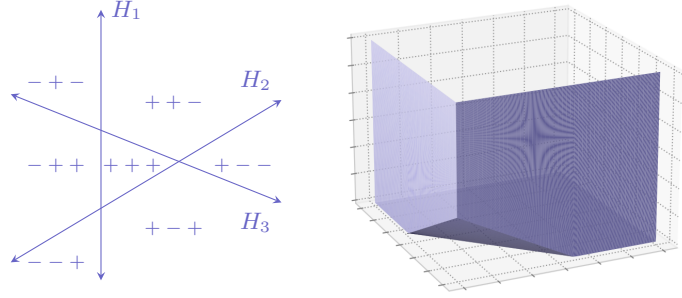
\begin{figure}[t]
\begin{tikzpicture}[>=stealth, scale=0.6]
    \draw[<->,cornflower!80!blue] (-1,3) --node[at start,right]{{\footnotesize $H_1$}} (-1,-3);
    \draw[<->,cornflower!80!blue] (-3,-2.6) --node[at end,above left]{{\footnotesize $H_2$}} (3,1);
    \draw[<->,cornflower!80!blue] (-3,1.128) --node[at end,below left]{{\footnotesize $H_3$}} (3, 1.128 - 2.4);

    \node[cornflower!80!blue] at (-0.31,-0.45) {{\tiny $+++$}};
    \node[cornflower!80!blue] at (-2.0,-0.45) {{\tiny $-++$}};
    \node[cornflower!80!blue] at (-2.0,1.4) {{\tiny $-+-$}};
    \node[cornflower!80!blue] at (-2.0,-2.6) {{\tiny $--+$}};
    \node[cornflower!80!blue] at (0.61,0.95) {{\tiny $++-$}};
    \node[cornflower!80!blue] at (0.61,-1.85) {{\tiny $+-+$}};
    \node[cornflower!80!blue] at (2.15,-0.45) {{\tiny $+--$}};
    
    \node[inner sep=0pt] (surface) at (8,0)
    {\includegraphics[width=0.35\textwidth]{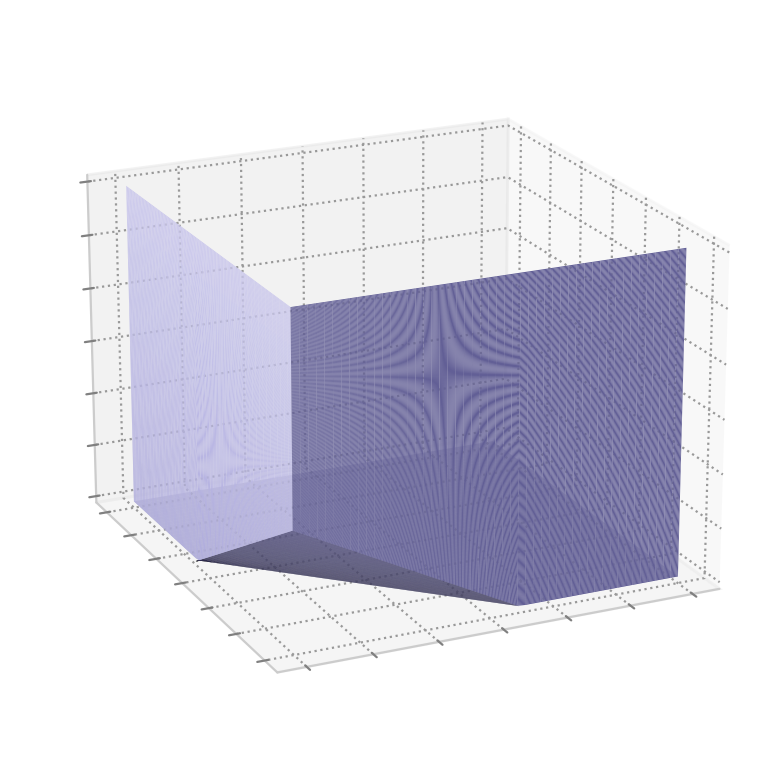}};
\end{tikzpicture}
\caption{Example hyperplane arrangement $\textbf{A}_1$ (left) and corresponding image of first layer map $F_1:\bb R^2\to\bb R^3$ (right), where $\theta\in C_0$.}
\label{fig:C0-config}
\end{figure}

As in Remark \ref{remark:codeword-and-intercepts}, in each configuration $P_+$ denotes the hyperplane in $\bb R^{n_0+1}$ that is the image of the first layer pre-activation map. We denote the intercept tuple of $P_+$ by ${\bf p} = (p_0,p_1,\ldots,p_{n_0+1})$, which is well-defined since $\textbf{A}_1$ is generic.  Then $P_+$ is the affine hull of $F_1(R_{+})$. For example, Figure \ref{fig:C0-config} depicts an example hyperplane arrangement $\textbf{A}_1(\theta)$ and image set $\ImF$ when $n_0=2$, $n_1=3$, and we have $\theta\in C_0$. The region that is in the positive half space of every hyperplane is indicated and its image in $\bb R^3$ is the shaded darkest in the figure.

\subsection{ Case 1: $C_0$}
\label{subsec:simplex-case}
When $\theta \in C_0$,  $R_{+}=B$, and $\ImF$ looks like a hyperplane, with normal vector in the positive orthant, that has been ``bent'' into the positive orthant. 
By Lemma \ref{lem:intercepts-signs}, each element of the intercept tuple $\bf p$ is positive.

Rather than calculate $\delta_0 = \P(E^+|\ \theta\in C_0)$ directly, we indicate a collection of hyperplanes so that $\delta_0$ can be understood via the probability of $H$ being in that collection. For ${\bf p}$ as above, consider the collection of hyperplanes $\mathcal H_{\bf p}$, first discussed in Section \ref{subsec:hyperplane-partition}. Furthermore, recall the scalars $\lambda_1, \lambda_2, \ldots, \lambda_{n_0+1}$ that determine a hyperplane in $\mathcal H_{\bf p}$. 

We will show that $\delta_0 = \frac12\bb P(H\in\mathcal{H}^1_{\bf p})$, where $\mathcal H^1_{\bf p}$ is the subset of $\mathcal H_{\bf p}$ defined as 
    \[\mathcal H^1_{\bf p} = \{H\in \mathcal H_{\bf p}\ |\ \lambda_i \le 1, i=1,2,\ldots,n_0+1\}.\]

First, let $H$ be a hyperplane in $\R^{n_0+1}$ with an intercept tuple ${\bf q}=(q_1,\ldots,q_{n_0+1})$. We claim that in Case 1, and in the event $E^+$, we have $\ImF \subset H^-$ if and only if $H \in \mathcal H^1_{\bf p}$ and $H$ has a positive bias {--} which is equal, in probability, to the condition $0 < q_i < p_i$, for all $i$, and $H$ has a positive bias (using Lemma \ref{lem:prob-mathcalP=0}).

To show the claim, note that an element of $\{+,-\}^{n_0+1}$ which has exactly one positive component, the $i^{th}$ component say, must be contained in $\operatorname{code}(\textbf{A}_1)$. It will be associated to some unbounded region, $R_i$. A similar argument to the one found in the proof of Lemma \ref{Lemma:ImF1}, shows that, for every $1\le i\le n_0+1$, the intersection $\ImF\cap\halfax{i}$ is unbounded. Letting ${\bf e}_i$ be the $i^{th}$ standard basis vector, the point $p_i{\bf e}_i$ is the image of the vertex $v_i$ of $\overline{B}$ that is in the boundary of $\overline{R}_i$. Since $F_1(\overline{R}_i)$ is convex (Remark \ref{remark:regions-convex-image}), it must be that for every $x_i \ge p_i$, the point $x_i{\bf e}_i$ is in $\ImF\cap\halfax{i}$. Thus, if $\ImF\cap H = \emptyset$ then $q_i < p_i$ for all $1\le i\le n_0+1$. 

By Remark \ref{rem:H-positiveorthant}, we must have that $H$ cuts through the positive orthant, so there is some $i$ with $q_i > 0$, almost surely. Suppose there were some $j$, $1\le j\le n_0+1$, so that $q_{j} < 0$. Then the intersection of $H$ with the plane $\mathtt{X}_{\{i,j\}}$ is a line with a positive slope. Since $P_+\cap\mathtt{X}_{\{i,j\}}$ has negative slope and $0 < q_i < p_i$, we see that $H$ and $P_+$ would necessarily intersect in $\halfax{\{i,j\}}$, giving a point in $H\cap \ImF$. Hence, if $\ImF\subset H^-$ then, almost surely, $0 < q_i < p_i$, meaning that $H\in\mathcal H^1_{\bf p}$.The bias for $H$ must be positive since $H$ separates any one of the points $p_i{\bf e}_i$ from the origin and $p_i{\bf e}_i\in H^-$. The other direction is clear, that is, if $H$ has a positive bias and $H\in\mathcal H^1_{\bf p}$ then $\ImF\subset H^-$. 

In summary, given that $\theta\in C_0$, the event $E^+$ has the same probability as the event that $H\in\mathcal H^1_{\bf p}$ and $H$ has a positive bias. Hence, $\delta_0 = \frac12\bb P(H\in\mathcal{H}^1_{\bf p})$.

\subsection{Case 2: $C_1,\ldots,C_{n_0+1}$}
\label{subsec:facet-case}

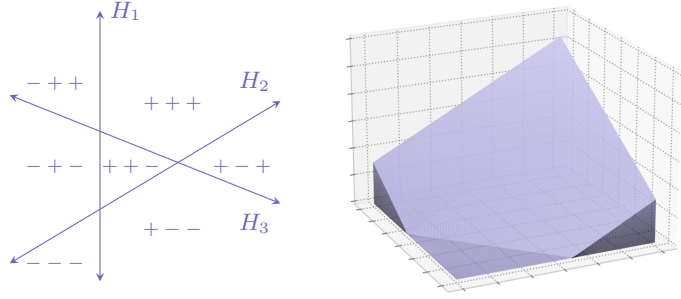
\begin{figure}[t]
\begin{tikzpicture}[>=stealth, scale=0.6]
    \draw[<->,cornflower!80!blue] (-1,3) --node[at start,right]{{\footnotesize $H_1$}} (-1,-3);
    \draw[<->,cornflower!80!blue] (-3,-2.6) --node[at end,above left]{{\footnotesize $H_2$}} (3,1);
    \draw[<->,cornflower!80!blue] (-3,1.128) --node[at end,below left]{{\footnotesize $H_3$}} (3, 1.128 - 2.4);

    \node[cornflower!80!blue] at (-0.31,-0.45) {{\tiny $++-$}};
    \node[cornflower!80!blue] at (-2.0,-0.45) {{\tiny $-+-$}};
    \node[cornflower!80!blue] at (-2.0,1.4) {{\tiny $-++$}};
    \node[cornflower!80!blue] at (-2.0,-2.6) {{\tiny $---$}};
    \node[cornflower!80!blue] at (0.61,0.95) {{\tiny $+++$}};
    \node[cornflower!80!blue] at (0.61,-1.85) {{\tiny $+--$}};
    \node[cornflower!80!blue] at (2.15,-0.45) {{\tiny $+-+$}};
    
    \node[inner sep=0pt] (surface) at (8,0)
    {\includegraphics[width=0.35\textwidth]{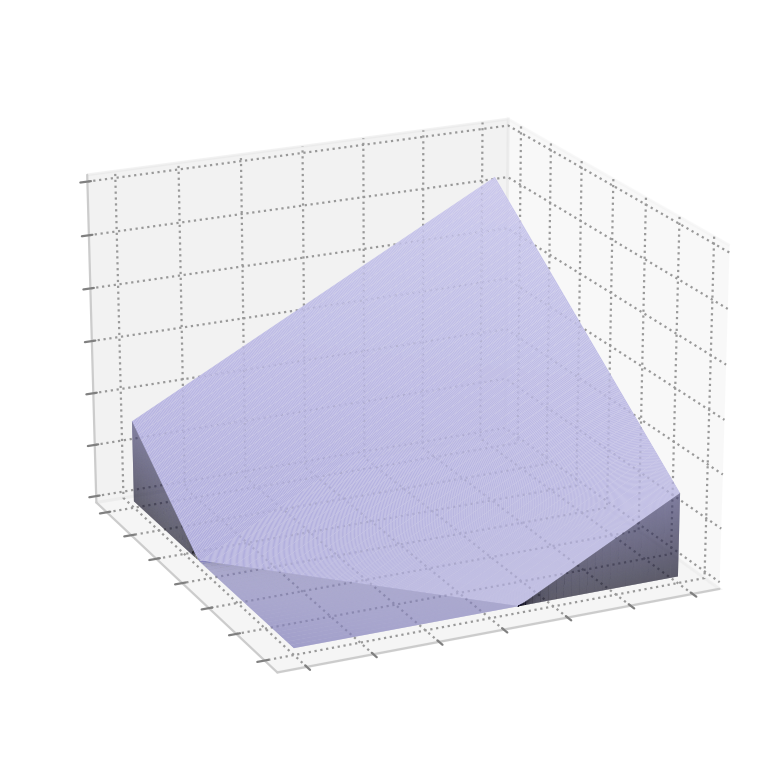}};
\end{tikzpicture}
\caption{Hyperplane arrangement $\textbf{A}_1$ (left) and image of first layer map $F_1:\bb R^2\to\bb R^3$ (right) for example $\theta\in C_3$; $n_0=2$, $n_1=3$.}
\label{fig:facet-config}
\end{figure}

When $\theta \in {C_i}$, $i=1, \ldots, n_0+1$, then $\overline{R}_{+}$ intersects $\overline{B}$ at a facet  and $\operatorname{code}(B)$ has exactly $1$ negative sign in the $i^{th}$ component (see Figure \ref{fig:facet-config} for the case $n_0=2$). By Lemma \ref{Lemma:ImF1} and its proof, $\halfax{i}\cap\ImF$ consists of just the origin and $\halfax{j}\subset\ImF$, for each $1\le j\le n_0+1$ with $j\ne i$.

In this case the hyperplane $P_+$ has an intercept tuple ${\bf p} = (p_0,p_1,\ldots, p_{n_0+1})$ which satisfies $p_i < 0$, and $p_j > 0$ for $j\ne i$, by Lemma \ref{lem:intercepts-signs}.

As with $\delta_0$, we do not find each of $\delta_1, \delta_2, \ldots, \delta_{n_0+1}$ individually. However, we can use the symmetries of the configurations and results on hyperplanes from Section \ref{sec:lemmas} to calculate the sum $\sum_{i=1}^{n_0+1}\delta_i$.

\begin{lem}
\label{lem:facet-hyperplanes}
Let the hyperplane $P_+$ and the $(n_0+1)$-tuple ${\bf p}$ be as in the setup above, and let $\mathcal{H}_{-{\bf p}}$ have partition $\{\mathcal P, \mathcal S_1,\ldots, \mathcal S_{n_0+1}\}$ as in Lemma \ref{lem: partition}. Then $H\in\mathcal H_{-\bf p}$ if the parameter $\theta$ is in the event $E^+$. Furthermore, $\P(E^+| \theta \in C_i) = \frac{1}{2}\P(H \in \mathcal{S}_i)$, for $i\in\{1,\ldots, n_0+1\}$.

\end{lem}
\begin{proof}
 Let $i$ be such that $\theta\in C_i$, with $1\le i\le n_0+1$. As in Lemma \ref{lem: partition}, we consider $\mathcal{H}_{\bf -p}$, the set of all hyperplanes with intercept tuples ${\bf q}=(q_1,\ldots,q_{n_0+1})$ where $q_j$ has the same sign as $-p_j$, for $j=1,2,\ldots,n_0+1$. From Lemma \ref{Lemma:ImF1}, we know that $\ImF$ contains $\halfax{j}$ for all $j \neq i$. Thus, we know that the set of hyperplanes in $E^+$ must be a subset of $\mathcal{H}_{\bf -p}$ since, almost surely, a hyperplane not in $\mathcal{H}_{\bf -p}$ either intersects $\halfax{j}$ away from the origin for some $j\ne i$ (recall that $p_j > 0$ if $j\ne i$) or, since $-p_i > 0$, it has the property that $q_j < 0$ for all $j=1,2,\ldots, n_0+1$ which, as we have remarked, cannot occur if the event $E^+$ occurs. 
 
 The slope of the intersection of $P_+$ with the coordinate plane $\mathtt{X}_{\{j,i\}}$ (oriented with $j^{th}$ component first and $i^{th}$ component second) is given by $m_{ji}=-\frac{p_i}{p_j}$.

Let $H' \in \mathcal{S}_k \subset \mathcal{H}_{\bf -p}$, where $k \neq i$, have intersection tuple ${\bf q'}=(q_1',\ldots,q_{n_0+1}')$. Then as in Lemma \ref{lem: partition}, $\lambda_k > \lambda_{k'}$ for all $k' \neq k$, where $\lambda_k$ is the scalar such that $q_k' = \lambda_k(-p_k)$. Then the slope of the intersection of $H'$ with the plane $\mathtt{X}_{\{k,i\}}$ is given by $m'_{ki}=-\frac{q_i'}{q_k'}=-\frac{\lambda_i(-p_i)}{\lambda_k(-p_k)}$. On the other hand, the slope of the intersection of $P_+$ in the plane $\mathtt{X}_{\{k,i\}}$ is $-\frac{p_i}{p_k}$, which is positive. Since $\lambda_k > \lambda_i$, $\frac{\lambda_i}{\lambda_k} < 1$, and so

\[-\frac{\lambda_i(-p_i)}{\lambda_k(-p_k)} < -\frac{p_i}{p_k}.\]

Since $p_k$ is positive, and we know $q_k'$ and $-p_k$ have the same sign, $q_k' < 0 < p_k$. Thus, $H'$ has a smaller $k^{th}$ intercept than $P_+$, and a smaller (positive) slope in the $\mathtt{X}_{\{k,i\}}$ plane, and so $H'$ must intersect $P_+$ in the positive quadrant $\halfax{\{k,i\}}$. Therefore, when $\theta \in E^+ \cap C_i$
\[\P(\ImF \subset (H')^-| H' \in \mathcal{S}_k) = 0, \quad \forall k \neq i.\]

Using similar reasoning, one can show that any $H'\in\mathcal{S}_i$ does not intersect $P_+$ in $\halfax{\{j,i\}}$ for any $j\ne i$. Thus, with probability 1, $H'$ does not intersect $P_+$ in the positive orthant.\footnote{If $H'$ and $P_+$ have non-empty intersection then, generically, the intersection is an $(n_0-1)$-dimensional affine subspace in $\bb R^{n_0+1}$ and that affine subspace would intersect any plane $\mathtt{X}_{\{j,i\}}$ in a point.} 

From our reasoning above, except for the possibility that $H\in\mathcal P$, we have that $\ImF \subset H^-$ if and only if $H \in \mathcal{S}_i$ and $H$ has a negative bias (the origin is in $\ImF$ in this case, so it needs be in $H^-$). From Lemma \ref{lem:prob-mathcalP=0}, $\P(H \in \mathcal{P})= 0$. Therefore, 
\[\P(E^+ | \theta \in C_i) = \frac{1}{2}\P(H \in \mathcal{S}_i\cup\mathcal{P})= \frac{1}{2}\P(H \in \mathcal{S}_i).\]
\end{proof}

With Lemma \ref{lem:facet-hyperplanes} in hand, we are able to prove that 
    \[\sum_{i=1}^{n_0+1}\delta_i=\frac{1}{2^{n_0+2}}.\]
To do so, consider any intercept tuple ${\bf p}=(p_1, \ldots, p_{n_0+1})\in\R^{n_0+1}$, let $\{\mathcal{S}_1,\ldots,\mathcal{S}_{n_0+1},\mathcal{P}\}$ be the partition of $\mathcal{H}_{\bf p}$ from Lemma \ref{lem: partition}.
A useful property of this partition is that the size of each of $\mathcal{P}, \mathcal{S}_1,\ldots, \mathcal{S}_{n_0+1}$ only depends on the magnitude of each component in $\bf p$, not the sign. Indeed,
let $H \in \mathcal{H}_{\bf p}$ be a hyperplane with intersection tuple ${\bf q}=(q_1,\ldots,q_{n_0+1})$ and, for some $\epsilon=(\epsilon_1,\ldots,\epsilon_{n_0+1})\in\{1,-1\}^{n_0+1}$, consider $H' \in \mathcal{H}_{\epsilon\odot\bf p}$ to be the hyperplane with intersection tuple $\epsilon\odot{\bf q}$, where $\odot$ is the Hadamard product (e.g., the $j^{th}$ component of $\epsilon\odot{\bf p}$ is $\epsilon_jp_j$). For each $i$, as previously, write $\lambda_i$ for the scalar such that $q_i = \lambda_i p_i$. We have $H \in \mathcal{S}_i$ if and only if $\lambda_i > \lambda_j$, for all $j \neq i$ and the scalars $\lambda_1,\ldots,\lambda_{n_0+1}$ are pairwise distinct. Note that $\epsilon_jq_j = \lambda_j\epsilon_jp_j$ for every $j=1,2,\ldots,n_0+1$ and so $H'$ is determined in $\Hp{\epsilon\odot{\bf p}}$ by the same scalars $\lambda_1,\lambda_2,\ldots,\lambda_{n_0+1}$. Letting $\mathcal{S}'_1,\ldots,\mathcal{S}'_{n_0+1},\mathcal{P}'$ be the partition of $\mathcal{H}_{\epsilon\odot{\bf p}}$, we see that $H\in\mathcal{S}_i$ if and only if $H'\in\mathcal{S}'_i$. 

Recall the discussion from Section \ref{sec:lemmas} of how the intercepts of $P_+$ and the intercepts of $H$ are determined from the weights and biases. 
As the weights and biases are assumed to be identically and symmetrically distributed around the origin,
then the components of ${\bf q}$ and those of $\varepsilon\odot{\bf q}$ have identical distributions. Hence, for the hyperplane $H \in \mathcal H_{-\bf p}$ associated to a neuron in the second layer, and for a given $\epsilon\in\{1,-1\}^{n_0+1}$, we have that $\bb P(H\in\mathcal{S}_i) = \bb P(H\in\mathcal{S}'_i)$ (using the notation of the previous paragraph).

To compare different configurations on $\textbf{A}_1$ with the same underlying arrangement, changes in $(\operatorname{sign}(p_1), \operatorname{sign}(p_2), \ldots, \operatorname{sign}(p_{n_0+1}))$ produce all intercept tuples that arise from the cases $C_1,\ldots,C_{n_0+1}$ (see Remark \ref{remark:codeword-and-intercepts}).\footnote{By considering only those that make exactly one of the signs negative.} Combining this with Lemma \ref{lem:facet-hyperplanes}, 
we have found that we may simply consider ${\bf p}$ to be determined from some $\theta\in C_1$, say, in order to compute the sum of $\delta_1,\delta_2,\ldots,\delta_{n_0+1}$. Letting the partition of $\mathcal H_{-\bf p}$ be $\{\mathcal{S}_1,\ldots,\mathcal{S}_{n_0+1},\mathcal{P}\}$, then 
\[\sum_{i=1}^{n_0+1}\delta_i = \frac12 \sum_{i=1}^{n_0+1}\bb P(H\in \mathcal{S}_i) = \frac12 \bb P(H\in \mathcal H_{-\bf p}).\]

From our assumption that weights and the bias for $H$ are independently and symmetrically distributed about 0, it is clear that each $q_i$ in the intercept tuple of $H$ has a $1/2$ probability of being positive. Thus, it is clear that $\P(H\in \mathcal{H}_{-\bf p})=\frac{1}{2^{n_0+1}}$. We have, therefore, found that 
\[\sum_{i=1}^{n_0+1}\delta_i=\frac{1}{2^{n_0+2}}.\]

\subsection{Case 3: $C_{n_0+2},\ldots, C_{2n_0+2}$}
\label{subsec:vertex-case}

When $\theta \in C_i$, $i=n_0+2, \ldots,  2n_0+2$, then  $\overline{R}_+$ intersects $\overline{B}$ at a vertex and $\operatorname{code}(B)$ has only one positive component in the $(i-n_0-1)^{th}$ component; so $B$ is in the positive half-space of only the $(i-n_0-1)^{th}$ hyperplane in $\textbf{A}_1$. As discussed in Lemma \ref{Lemma:ImF1} and its proof, $F_1(B)$ is contained in a bounded portion of the coordinate axis associated with that hyperplane. Furthermore, for an index $j$, with $j\ne i-(n_0+1)$, we have $\halfax{j}\subset\ImF$.

By definition, $\theta \in C_i$ means that $\code(B)$ has the exact opposite signs compared to the codeword of the bounded region when a parameter is in $C_{i-n_0-1}$. Let $\bf p$ be the intersection tuple for the hyperplane $P_+$, as above, when $\theta \in C_i, i=n_0+2,\ldots,2n_0+2$. We negate every weight and bias associated to the first layer in $\theta$, getting a new parameter $\theta'$. Let $P'_+$ be the hyperplane that is likewise determined by $\operatorname{Im}(F^{\theta'}_1)$, and let ${\bf p}'$ be its corresponding intersection tuple. We have that $\theta' \in C_{i-n_0-1}$ and ${\bf p}=-{\bf p}'$.

Since $\ImF$ contains every $\halfax{j}$ that is a subset of $\operatorname{Im}(F^{\theta'}_1)$ (see Lemma \ref{Lemma:ImF1}(b)), if $H \not\in \Hp{-{\bf p}'}=\Hp{\bf p}$ then $H$ must either intersect $\ImF$ or have an intercept tuple $\bf q$ such that $q_j < 0$ for all $1\le j\le n_0+1$. As this would keep it from being in $E^+$, a hyperplane $H$ in the event $E^+$ must be in $\Hp{\bf p}$, given that $\theta \in C_i$. 

The slope of the intersection of $P_+$ in the coordinate plane $\mathtt{X}_{\{j,i\}}$ is given by $-\frac{p_i}{p_j}$, which is the same as the slope of the intersection of $P'_+$ with $\mathtt{X}_{\{j,i\}}$. Thus, we can use the same reasoning as in Lemma \ref{lem:facet-hyperplanes} to show that $H \in \Su_{i-n_0-1} \subset \Hp{\bf p}$ is necessary in order to have $\ImF\cap H = \emptyset$. While $H\in \Su_{i-n_0-1}$ is sufficient to have $\operatorname{Im}(F^{\theta'}_1)\cap H = \emptyset$, it does not imply $\ImF\cap H = \emptyset$.

Since the origin is contained in $\ImF$, a connected set, the cooriented hyperplanes that satisfy $E^+$ when $\theta \in C_i, i=n_0+2,\ldots,2n_0+2$ are precisely those which have a negative bias and also satisfy $\ImF\cap H = \emptyset$. And so we have that these hyperplanes are precisely those with correct coorientation in the set 
\[\Su_{i-n_0-1} \setminus \{H|  \ImF\cap H\ne\emptyset\}.\]

Recall the definition of $\Hp{\bf p}^{1}$, where $Q\in\Hp{\bf p}^{1}$ if and only if $Q\in\Hp{\bf p}$ and $\lambda_j \le 1$ for all $j$ (and $\lambda_1,\ldots,\lambda_{n_0+1}$ are the scalars for $Q\in\Hp{\bf p}$). Now, suppose that $H\in \Su_{i-n_0-1}$ is a hyperplane such that $\ImF\cap H\ne\emptyset$ and write $(q_1,q_2,\ldots,q_{n_0+1})$ for its intercept tuple. We have a unique intercept that is positive, namely $q_{i-n_0-1} > 0$, as $H\in\Hp{\bf p}$ and the configuration in this case requires $p_{i-n_0-1}$ to be the only positive element of ${\bf p}$. For simplicity of notation, write $i' = i - n_0 - 1$.

Suppose it were the case that $q_{i'} > p_{i'}$. Then, for all $j\ne i'$, the intersection of $H$ with any coordinate plane $\mathtt{X}_{\{j,i'\}}$ {--} using orientation so the $j^{th}$ coordinate is first {--} has slope $\frac{q_{i'}}{|q_j|}$; this is larger than $\frac{p_{i'}}{|p_j|}$, which is the slope of the intersection of $P_+$ with $\mathtt{X}_{\{j,i'\}}$. With $q_{i'} > p_{i'}$, this makes it impossible to have a point in $\ImF\cap H$. Hence, the assumption that $\ImF\cap H \ne \emptyset$ tells us that $q_{i'}=\lambda_{i'}p_{i'} \le p_{i'}$. Now, $H\in\Su_{i'}$ implies that $\lambda_j \le \lambda_{i'} \le 1$ for all $j=1,2,\ldots,n_0+1$, showing that $H\in\Hp{\bf p}^{1}$. Moreover, if $H\in\Hp{\bf p}^1$ then we must have $\ImF\cap H\ne\emptyset$ (the sets intersect along $\halfax{i'}$). 

Setting $\Su_j^{1} = \Su_j\cap\Hp{\bf p}^{1}$ for each $j=1,2,\ldots,n_0+1$, we have that all but a probability zero subset of hyperplanes in $\Hp{\bf p}^{1}$ are in $\Su^{1}_1 \cup \Su^1_2\ldots\cup \Su^{1}_{n_1}$. We have shown that if $\theta\in C_i$, with $n_0+2\le i\le 2n_0+2$, and $\theta'\in C_{i-n_0-1}$ is the related parameter defined above, then if a hyperplane $H\in \Su_{i-n_0-1}$ is such that $\ImF\cap H\ne\emptyset$, then, almost surely, $H\in\Su^1_{i-n_0-1}\subset\mathcal{H}^1_{\bf p}$, where ${\bf p}$ is the intercept tuple for$P_+$. 

By the same reasoning as in subsection \ref{subsec:facet-case}, while this statement about set containment requires a different tuple ${\bf p}$ for each $i=n_0+2,\ldots,2n_0+2$, for the sake of the probability of $E^+$, any ${\bf p}$ with a single positive component suffices. Using the observation above that a negative bias is needed, we have that for any $n_0+2\le i\le 2n_0+2$, 
\[\bb P(E^+\ |\ \theta\in C_{i-n_0-1}) - \bb P(E^+\ |\ \theta\in C_{i}) = \frac12\bb P(H \in \Su^1_{i-n_0-1}),\] where $H$ is the hyperplane given by $({\bf w}|b)$ which comprises part of $\theta$, and the indexed set $\Su_{i-n_0-1}$ is in the partition of $\Hp{\bf p}$, as in Lemma \ref{lem: partition}. Shifting indices, 
\[\sum_{i=1}^{n_0+1}(\delta_i - \delta_{i+{n_0+1}}) = 
\sum_{i=1}^{n_0+1}\frac12\bb P(H\in\Su_i^{1}) = \frac12\bb P(H \in \Hp{\bf p}^{1}) = \delta_0.\]

By Remark \ref{rem:zero-deltas} and findings in subsections \ref{subsec:simplex-case} through \ref{subsec:vertex-case}, we have 
    \begin{align*}
        \sum_{i=0}^{2^{n_1}-1} \delta_i 
        = \delta_0 + 2\sum_{i=1}^{n_0+1}\delta_i + \sum_{i=1}^{n_0+1}(\delta_{i+n_0+1} - \delta_i) 
        = \delta_0 + 2\frac{1}{2^{n_0+2}} - \delta_0 
        = \frac{1}{2^{n_0+1}}
    \end{align*}
which concludes the proof of  Lemma \ref{Lemma: Sum of delta}.

\section{Other architectures}\label{sec:general architectures}
In this section we discuss Conjecture \ref{conj:large-n1}. Throughout the section, we consider neural networks $\mathcal N$ with architecture $(n_0,n_1,\ldots)$ such that $n_0$ is fixed and $n_1 > n_0$. Conjecture \ref{conj:large-n1} states that there is a constant $c > 0$ such that, for $n_1$ sufficiently large, the probability of a given neuron in the second layer of $\mathcal N$ being stably unactivated is at least $c\frac{1}{4^{n_0}}$. Below we give some empirical evidence of this conjecture, as well as some rationale that it holds generally. Note that this conjecture, if true, would imply that for $n_1$ sufficiently large $\bb P(E) \approx \bb P(E^+)$, since it is necessarily the case that $\bb P(E^-) = \frac{1}{2^{n_1+1}}$.

\subsection{Computational evidence}
In support of the Conjecture \ref{conj:large-n1}, we present empirical results on the value of $\bb P(E)$  when $2\le n_0\le 7$, for values of $n_1$ satisfying $n_0+2 \le n_1 \le n_0+16$. For each pair $(n_0, n_1)$ in that range, we sampled 100,000 parameters $\theta$ for a ReLU network with architecture $(n_0, n_1, 1)$, applying the He-uniform initialization to both weights and biases \cite{HeZhangRenSun2015}. 
We tested whether a neuron is stably unactivated using a Monte Carlo method,
sampling the domain to test for only negative values at the given neuron. A computational challenge with this approach lies in the density of the sampling; small regions in the domain where the values could be positive might be missed by the sampling. As a result, only testing at $\theta$ would result in a significant number of false positives (returning that the neuron is stably unactivated when it is not), while false negatives never occur with that approach. 

Considering Definition \ref{defn:stably-unactivated}, we attempted to address this shortcoming as follows. For each parameter $\theta$, we randomly selected a set of parameters $\theta_1,\ldots,\theta_{4n_0}$ from a small neighborhood in $\Omega$ centered at $\theta$. The values of the given neuron were then tested at each of these nearby parameters within the domain sample. This creates a possibility of returning a false negative since one of the selected nearby parameters  might lie outside of a neighborhood $O$ that satisfies Definition \ref{defn:stably-unactivated}. Greater accuracy of the results may then be obtained, having both false positives and false negatives, despite the computational constraints.

\begin{figure}[t]
    \includegraphics[width=\textwidth]{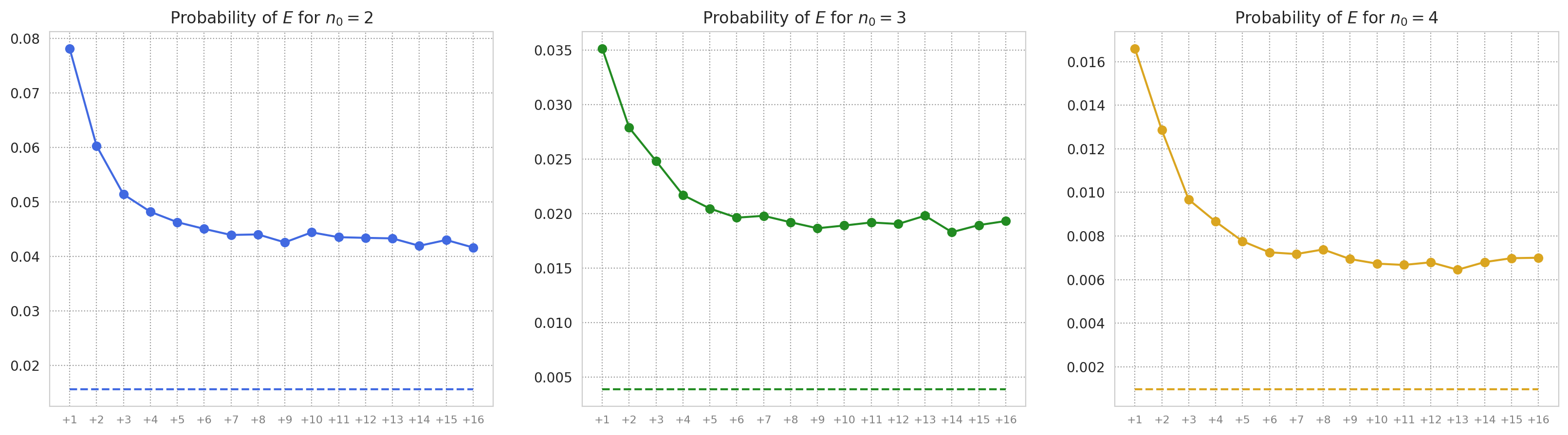}
    \caption{Empirical probabilities of a neuron in the second layer being stably unactivated for architectures with $n_0 = 2, 3,$ and $4$. Along the horizontal axis we show the amount by which $n_1$ is larger than $n_0$; the dashed horizontal line is at height $1/4^{n_0+1}$.}
    \label{fig:data2-4}
\end{figure}

\begin{figure}[h]
    \includegraphics[width=\textwidth]{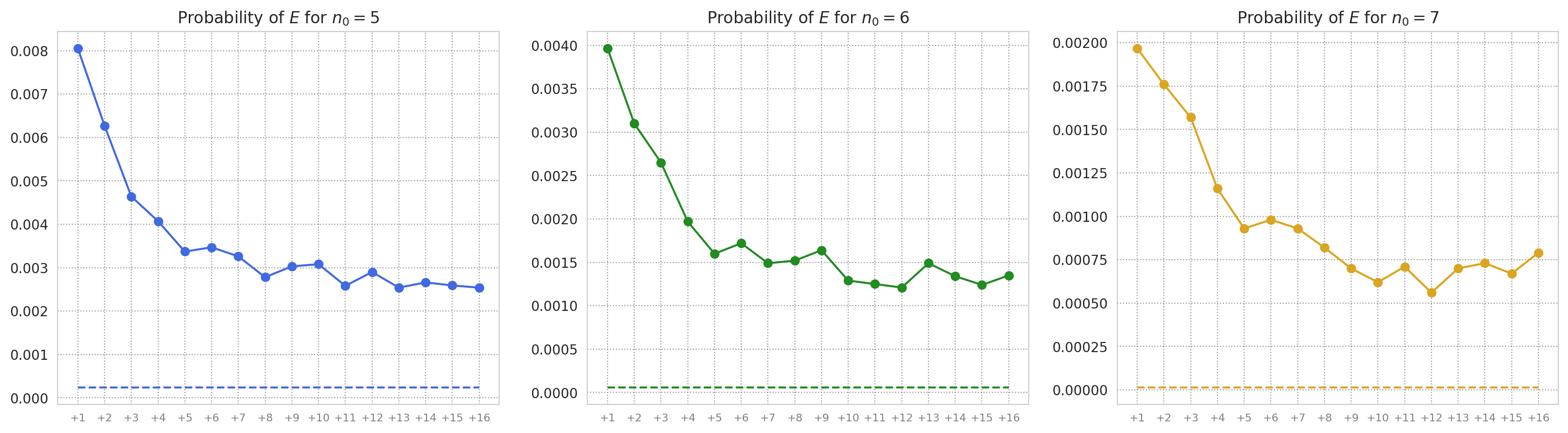}
    \caption{Empirical probabilities of a neuron in the second layer being stably unactivated for architectures with $n_0 = 5, 6,$ and $7$. Along the horizontal axis we show the amount by which $n_1$ is larger than $n_0$; the dashed horizontal line is at height $1/4^{n_0+1}$.}
    \label{fig:data5-7}
\end{figure}

Our results are plotted in Figure \ref{fig:data2-4} and Figure \ref{fig:data5-7}. In these plots, we include the value of $\bb P(E)$ when $n_1=n_0+1$ that is known from Theorem \ref{thm:main}. Each of the given curves indicates that the ratio of the empirical probabilities to $1/4^{n_0}$ does not decrease as $n_0$ grows. In fact, we observe an increasing ratio, which may indicate that the bound in Conjecture \ref{conj:large-n1} could be tightened.

\subsection{Rationale for the Conjecture}
As in Sections \ref{sec:proof-main} and \ref{sec:configs}, we assume the parameter $\theta$ for the network is such that $\textbf{A}_1 = \textbf{A}_1(\theta)$ is a generic hyperplane arrangement; moreover, as above we simply write $F_1$ to denote the first layer map $F^{\theta}_1:\bb R^{n_0}\to\bb R^{n_1}$, and $H$ refers to the cooriented hyperplane in $\bb R^{n_1}$ determined by the weights and bias for the given neuron in the second layer of $\mathcal N$. Moreover, as above we assume that weights and biases in each layer are continuous i.i.d.\ random variables with a distribution that is symmetric about $0$.

To begin, we remark that the probability of $E$ decreases as a function of $n_1$, under typical assumptions on the underlying distribution for the weights and biases. Indeed, consider $F_1$ and $H$ as above, corresponding to a network with architecture $(n_0, m, \ldots)$. Let $F'_1$ be the first layer map of a network with the same values of $n_0$, but with $n_1 = m+1$, and let $H'$ be a random hyperplane in $\bb R^{m+1}$ determined by $m+1$ weights and a bias. Assume that there is a positive scalar so that, as random variables, the weights and bias for the hyperplane $H'$ are equivalent to that scalar times the weights and biases determining $H$.\footnote{This is a common situation. For example, if the ReLU networks are initialized with the He-uniform or He-normal distribution \cite{HeZhangRenSun2015}, then increasing $n_1$ by one will be equivalent to scaling all weights and biases in the second layer by a constant determined by $n_1$.} Since the hyperplane distribution for $H$ is invariant under overall scaling (importantly, the biases are \emph{not} initialized as zero), we have that $H'$ and $H$ have equal distributions. 

Write $H'_0$ for the projection of the intersection of $H'$ with the coordinate hyperplane $\mathtt{X}_{\{1,\ldots,m\}}$ to its first $m$ coordinates; $H'_0\subset \bb R^m$ inherits a coorientation from that of $H'$. Additionally, use the notation $\operatorname{Im}(F_1')_0$ for the set defined in the analagous way from $\operatorname{Im}(F_1')$. Now, clearly if $\operatorname{Im}(F'_1)\subset (H')^-$ then $\operatorname{Im}(F'_1)_0 \subset (H'_0)^-$. However, by our assumptions on the weights and biases, $\bb P\left(\operatorname{Im}(F'_1)_0 \subset (H'_0)^-\right) = \bb P(\ImF \subset H^-)$, which implies that $\bb P(\operatorname{Im}(F'_1)\subset (H')^-) \le \bb P(\ImF \subset H^-)$.

As in Section \ref{sec:configs}, use $P_+$ to denote the image of the first layer pre-activation map of $\mathcal N$. With our assumption of a generic $\textbf{A}_1$, $P_+$ is an $n_0$-dimensional affine subspace of $\bb R^{n_1}$. In Section \ref{sec:configs}, since we had $n_1 = n_0+1$, this is a hyperplane and it was beneficial to describe $P_+$ in terms of its intercept tuple. Moreover, we used in this case that a hyperplane that does not intersect $P_+$ would necessarily be parallel {--} with an intercept tuple that is one multiple of that of $P_+$. Those hyperplanes which do not intersect $P_+\cap\bb R^{n_1}_{\ge0}$ were described by using a partition of $\Hp{\bf p}$ which separated hyperplanes according to intervals in which the components of their intercepts tuples lied.

However, in the present context the generic assumption does not provide an intercept tuple for $P_+$. We have that $\dim P_+ = n_0$. Choosing a hyperplane $H_+$ in $\bb R^{n_1}$ which satisfies $P_+\subset H_+$, we have that $H_+$ may be chosen from an $(n_1-n_0-1)$-dimensional space of hyperplanes which contain $P_+$. Any non-zero translation in its normal direction of such an $H_+$ is a hyperplane that has empty intersection with $P_+$. Thus, for an $H$ that does not intersect $P_+$, we have a freedom of choice for $n_1 - n_0 - 1$ of the coordinate axis intercepts of $H$ with the remaining $n_0+1$ of the intercepts fixed. And so, for the hyperplane to have empty intersection with $P_+\cap\bb R_{\ge0}$, we should then expect that there is a subset of hyperplanes whose intercepts in $n_0+1$ of the coordinates are restricted to some interval, but are not restricted in the other coordinate directions. 

By Lemmas \ref{lem:bounded-dominates} and \ref{lem:average-number-of-facets}, which we prove below, it can be expected that most of the coordinate axes will have compact intersection with $\ImF$.  
In fact, we check that, on average, less than $2n_0+1$ of the $n_1$ coordinate axes intersect $\ImF$ anywhere except, possibly, the origin (see Corollary \ref{cor:average-axis-intersections}).

Now, suppose we fix the coorientations of $n_0+1$ of the $n_1$ hyperplanes in $\textbf{A}_1$. We discussed in Section \ref{sec:configs} hyperplanes in $\bb R^{n_0+1}$ that will have empty intersection with the slice of $\ImF$ that is zero in all but these $n_0+1$ coordinates. In the remaining coordinates, the above discussion hints at the ability to allow (at least on average) for the intercepts of $H$, in those remaining coordinates, to be unrestrained, positive or negative {--} except, possibly, needing to avoid some interval in around $2n_0$ of them. 

There are $2^{n_1 - n_0 - 1}$ coorientations of $\textbf{A}_1$ which restrict to our one fixed choice of coorientation on these $n_0+1$ hyperplanes. Also, by the reasons above and relying on the work to prove Theorem \ref{thm:main}, we would guess that there is some constant $c > 0$ so that if we put together those conditional probabilities of $E^+$ (conditioned on the coorientation of $\textbf{A}_1$), where the coorientation on the unselected $n_1-n_0-1$ hyperplanes is fixed, then these probabilities sum to at least $c \frac{1}{2^{n_0+1}}$. Since each coorientation of the underlying hyperplane arrangement of $\textbf{A}_1$ has equal likelihood, this results in an estimate of the probability of $E^+$ being at least
    \[\frac1{2^{n_1}}2^{n_1 - n_0 - 1}\left(c \frac{1}{2^{n_0+1}}\right) = \frac{c}{4}\frac1{4^{n_0}}.\]

We end the section by proving Lemmas \ref{lem:bounded-dominates} and \ref{lem:average-number-of-facets} that are referenced above, as well as Corollary \ref{cor:average-axis-intersections}.

\begin{lem}
\label{lem:bounded-dominates}
    For large $n_1$, the number of bounded regions of $\textbf{A}_1$ dominates the number of unbounded regions. 
\end{lem}

\begin{proof}
    As a direct consequence of Lemma \ref{lem:more-hyperplanes-than-n}, we have that for $n_1 = m$, the number of bounded regions $b(\textbf{A}_1) = \binom{m-1}{n_0}$ is asymptotically $O(m^{n_0})$. On the other hand, asymptotically the number of unbounded regions is 
    \begin{align*}r(\textbf{A}_1) - b(\textbf{A}_1)
     &= \binom{m}{n_0} - \binom{m-1}{n_0} + \binom{m}{n_0-1} + O(m^{n_0-2}) \\ 
     &= \binom{m-1}{n_0-1}+\binom{m}{n_0-1} + O(m^{n_0-2}), 
    \end{align*}
    which is $O(m^{n_0-1})$.
\end{proof}

Recall Lemma \ref{Lemma:ImF1}(b), which says that, in the case $n_1 = n_0+1$, there can be at most one $j$ between $1$ and $n_1$ with $\halfax{j}\not\subset\ImF$, regardless of the configuration in which $\theta$ is found. Additionally, this lemma says that $2^{n_1} - 2n_1$ of the configurations $C_i$ are such that $\halfax{j}\subset\ImF$ for all $j$, which has the effect that $\delta_i = \bb P(E^+\ |\ \theta\in C_i) = 0$ for all such configurations. In part, this result followed from the unicity of $B$, the bounded region of $\textbf{A}_1$, and that it shared a facet with $n_1$ unbounded regions. 

In general, for $n_1 > n_0$, the bounded regions of $\textbf{A}_1$ do not have a uniform number of facets. Moreover, how many regions (bounded or unbounded) of $\textbf{A}_1$ there are with a given number of facets is not the same for all generic hyperplane arrangements of $n_1$ hyperplanes in $\bb R^{n_0}$. However, we can understand the average number of facets among the regions of $\textbf{A}_1$. Doing so, in the interest of simpler notation, we write that $\textbf{A}_1$ has $m$ hyperplanes, instead of $n_1$.

\begin{lem}
    \label{lem:average-number-of-facets}
    For a generic hyperplane arrangement $\textbf{A}_1$ of $m$ hyperplanes in $\bb R^{n_0}$, the limit of the average number of facets of a region of $\textbf{A}_1$, as $m\to\infty$, is equal to $2n_0$.
\end{lem}
\begin{proof}
Let $s(\textbf{A}_1)$ denote the \emph{total} number of ($(n_0-1)$-dimensional) facets that appear in $\textbf{A}_1$. Since each is a facet for exactly two regions, the average number of facets of a region of $\textbf{A}_1$ is $\frac{2s(\textbf{A}_1)}{r(\textbf{A}_1)}$.

Suppose that the hyperplanes in $\textbf{A}_1$ are listed as $\textbf{A}_1 = \{H_1,H_2,\ldots, H_{m}\}$. For each $i$, $1\le i\le m$, recall the induced arrangement in $H_i$, from Definition \ref{defn:induced-arrangement}, denoted $\textbf{A}_{H_i}$. As $\textbf{A}_1$ is generic, it is clear that each induced arrangement $\textbf{A}_{H_i}$ is equivalent to a generic arrangement in $\bb R^{n_0-1}$ of $m-1$ hyperplanes. Furthermore, we may consider the closure of each region of $\textbf{A}_{H_i}$ as corresponding to one of the facets, of a region of $\textbf{A}_1$, that is a subset of $H_i$. Hence, we have that $s(\textbf{A}_1) = m \sum_{k=0}^{n_0-1}\binom{m-1}{k}$, and so the average number of facets of a region of $\textbf{A}_1$ is 
    \[\frac{2m \sum_{k=0}^{n_0-1}\binom{m-1}{k}}{\sum_{k=0}^{n_0}\binom{m}{k}}.\]

Considering the average number of facets of $\textbf{A}_1$ as a function of $m$, we compute that 
    \[\frac{2s(\textbf{A}_1)}{r(\textbf{A}_1)} = \frac{2\left(\frac{m^{n_0}}{(n_0-1)!} - \frac{1}{2(n_0-3)!}m^{n_0-1} + O(m^{n_0-2})\right)}{\frac{m^{n_0}}{n_0!} - \frac{n_0-3}{2(n_0-1)!}m^{n_0-1} + O(m^{n_0-2})}.\]
As a consequence, we find that $\lim_{m\to\infty}\frac{2s(\textbf{A}_1)}{r(\textbf{A}_1)} = 2n_0$.
\end{proof}

In fact, for a slight improvement, we can refine the result of Lemma \ref{lem:average-number-of-facets} and check that as a function of $m$ the average number of facets is increasing, at least for sufficiently large $m$. To do this, let $\textbf{A}_1$ be as above and let $\textbf{A}'_1$ be a generic hyperplane arrangement of $m+1$ hyperplanes in $\bb R^{n_0}$. By a lengthy, but straightforward, computation in which highest and second-highest order terms vanish, one can check that 
    \[\frac{s(\textbf{A}'_1)}{r(\textbf{A}'_1)} - \frac{s(\textbf{A}_1)}{r(\textbf{A}_1)} = \frac{m^{2n_0-2} + O(m^{2n_0-3})}{n_0!(n_0-1)!r(\textbf{A}'_1)r(\textbf{A}_1)},\]
and so the difference is positive for large enough $m$. Hence, for large $m$ the average number of facets is strictly less than $2n_0$.

\begin{cor}
\label{cor:average-axis-intersections}
    For sufficiently large $n_1$, the expected number of coordinate axes whose intersection with $\ImF$ consists of more than the origin is at most $2n_0+1$.
\end{cor}
\begin{proof}
Given a region $R$ of $\textbf{A}_1$, define $R$ to be \emph{\bf maximally negative} if there does not exist a region of $\textbf{A}_1$ whose codeword has more negative components than does $\operatorname{code}(R)$. Given that $R$ is a maximally negative region of $\textbf{A}_1$, say that $k$ is the number of facets of the closure $\overline{R}$ and that $H_1,H_2,\ldots, H_k$ are the hyperplanes in $\textbf{A}_1$ that are equal to the affine hulls of these facets. We remark that $R \subset H_1^{-}\cap H_2^{-}\cap\ldots\cap H_k^{-}$. To the contrary, if $R$ were in the positive half space of $H_i$, for some $1\le i\le k$, and $R_i$ is the region across that facet (so that $\overline{R}\cap\overline{R}_i$ is the facet in $H_i$), then $\operatorname{code}(R_i)$ and $\operatorname{code}(R)$ agree in all components but the $i^{th}$; yet, $\operatorname{code}(R)$ is positive in that component and $\operatorname{code}(R_i)$ is negative, contradicting that $R$ was maximally negative.

In addition, since all facets of $R$ arise from $H_1,H_2,\ldots, H_k$, a point in $\bb R^{n_0}$ is contained in $H_1^{-}\cap H_2^{-}\cap\ldots\cap H_k^{-}$ only if it is in $R$, which means that $R$ is the \emph{only} region of $\textbf{A}_1$ whose codeword is negative in components $1,2,\ldots, k$. Any region of $\textbf{A}_1$ which is mapped by $F_1$ into some coordinate axis must have a codeword with exactly one positive component. By what we have just observed, this requires either the positive component to be one of the first $k$ components or the region must be $R$ itself. Hence, if a maximally negative region has $k$ facets then at most $k+1$ regions of $\textbf{A}_1$ can be mapped to a subset of $\halfax{j}$ for some $1\le j\le n_1$.

Our assumptions make each coorientation of the arrangement underlying $\textbf{A}_1$ be equally likely, which makes the expected number of facets of a maximally negative region of $\textbf{A}_1$ equal to the average number of facets. Hence, for sufficiently large $n_1$, the expected number of $j$ such that $\halfax{j}\cap\ImF$ consists of more than the origin is at most $2n_0+1$, using Lemma \ref{lem:average-number-of-facets} and the comments immediately following its proof.
\end{proof}

\vspace{12pt}

\subsection*{Acknowledgments} This work is the result of an REU project that was completed at Towson University in the summer of 2024. The REU was financially supported by the National Science Foundation, under grant DMS-2149865. Additionally, the authors wish to acknowledge the generous support from the TU Department of Mathematics and the Fisher College of Science and Mathematics.

\bibliographystyle{amsalpha}
\bibliography{refs}

\end{document}